\documentclass{ws-ijufks}
\graphicspath{{../belief_noisy_or_revise/fig/}}
\usepackage[super,sort&compress]{natbib}
\renewcommand{\citep}{\cite}
\newcounter{Exe}
\renewcommand{\theExe}{\arabic{Exe}}
\newcommand{\Exe}{\text{Example} \refstepcounter{Exe}\textbf{%
            \theExe}}
\begin{document}

\markboth{Authors' Names}{Instructions for
Typesetting Camera-Ready Manuscripts}

\catchline{}{}{}{}{}
%
%%%%%%%%%%%%%%%%%%%%%%%%%%%%%%%%%%%%%%%%%%%%%%%%%%%%%%%%%%%%%%%%%%%%

\title{THE BELIEF NOISY-OR MODEL APPLIED TO NETWORK RELIABILITY ANALYSIS}

\author{KUANG ZHOU}

\address{Northwestern Polytechnical University\\
Xi'an, Shaanxi 710072, China\\~\\
DRUID, IRISA, University of Rennes 1\\ Rue E. Branly, 22300 Lannion, France\\
kzhoumath@163.com}

\author{ARNAUD MARTIN}

\address{DRUID, IRISA, University of Rennes 1\\ Rue E. Branly, 22300 Lannion, France\\
arnaud.martin@univ-rennes1.fr}

\author{QUAN PAN}

\address{Northwestern Polytechnical University\\
Xi'an, Shaanxi 710072, China\\
quanpan@nwpu.edu.cn}

\maketitle

\begin{history}
\received{(received date)}
\revised{(revised date)}
%\accepted{(Day Month Year)}
%\comby{(xxxxxxxxxx)}
\end{history}

\begin{abstract}
One difficulty  faced in knowledge engineering for
Bayesian Network (BN) is the quantification step where  the Conditional Probability Tables (CPTs) are determined. The number
of parameters  included in CPTs increases exponentially with the number of parent variables. The
most common solution is the application of the so-called
canonical gates. The Noisy-OR (NOR)  gate, which takes advantage of the
independence of causal interactions, provides a logarithmic
reduction of the number of parameters required to specify a CPT. In this paper, an extension of NOR model based on the theory of belief functions, named Belief Noisy-OR (BNOR),  is proposed. BNOR  is capable of dealing with both aleatory and epistemic uncertainty of the network.
Compared with NOR, more rich information which is of great value for making decisions can be got when the available
knowledge is uncertain. Specially, when there is no epistemic uncertainty, BNOR  degrades into NOR. Additionally, different structures of BNOR are presented in this paper in order to meet various needs of engineers.
The application of BNOR model on the reliability evaluation problem of networked systems demonstrates its effectiveness.
\end{abstract}

\setcounter{footnote}{0}
\renewcommand{\thefootnote}{\arabic{footnote}}

\section{Introduction}
Bayesian Network (BN) is a probabilistic graphical model  that represents a set of
random variables and their conditional dependencies via a directed acyclic graph (DAG) \cite{jensen1996introduction}.
BN can be used to learn causal relationships and  gain understanding
of a problem domain. It allows probabilistic beliefs to be updated automatically when new information becomes available.
BN is also able to represent multi-attribute correlated variables and to
perform relevant simulations or diagnoses. Owing these advantages, it has been widely applied on the problem of reliability or safety
analysis for both static and dynamic systems \citep{BNre1,BNre2,BNre3,BNre4}.
%a compact representation of a multivariate
%statistical distribution function and now is widely applied
%in reliability analysis for static or dynamic systems \citep{BNre1,BNre2,BNre3,BNre4}.
%\citet{bnsolution} presented a method for reliability evaluating
%of networks with imperfect nodes based on BN. See \citet{langseth2007bayesian}
%and \citet{weber2012overview} for overviews on  BN applied in
%reliability analysis problems.

In BN, Conditional Probability Tables (CPTs) should be defined to measure the relationships between variables. However, it has been pointed
out that it is usually difficult to  quantify the CPTs due to the complexity \cite{antonucci2011imprecise}.
One of the most appropriate solutions to this problem is the
Noisy-OR (NOR) gate, which can be attributed to \citet{pearl1988probabilistic}. Traditional
NOR can only deal with the binary variables. \citet{srinivas1993generalization}
extended NOR for $n$-ary input and output
variables, and  arbitrary functions other than Boolean OR function can be used.
But it has not taken into account the uncertainty on  parameters
and the state which  often exists in practice. Considering such uncertainty, \citet{fallet2012evidential} proposed the imprecise extensions of Noisy OR (ImNOR) thereafter.
Nevertheless, there are still some problems for
ImNOR which will be discussed in detail later.

The theory of belief functions, also called Dempster--Shafer Theory (DST), offers a mathematical framework for modeling uncertainty and imprecise
information \cite{ds2}. Belief functions are widely employed in various fields,
such as data classification \cite{denoeux1995k,liu2014credal,liu2015new},
data clustering \cite{masson2008ecm,denoeux2004evclus,zhou2015evidential, zhou2016ecmdd},
social network analysis \cite{wei2013identifying,zhou2014evidential,zhou2015median,zhou2015similarity}
    and statistical estimation \cite{denoeux2013maximum,come2009learning,zhou2014evidentialem}.
The concept of evidential networks,  which is  a combination of belief function theory
and Bayesian network, is proposed to  model system reliability with imprecise knowledge \cite{simon2008bayesian, simon2009evidential}. Recently, \citet{yaghlane2008inference} presented another definition of  evidential networks
based on  Transferable Belief Model (TBM) \cite{TBM}, Dempster-Shafer
rule of combination, and binary joint trees.

The objective of this work is to  enrich the existing NOR structures by integrating several
types of uncertainty. Under the framework of belief functions, the Belief Noisy-OR (BNOR) model is put forward. The uncertainty of variable states can be expressed by the power set of discernment frame. The uncertainty of parameters is described by probability intervals based on which
the basic belief assignments are determined.
The model can model causal connections among variables as well as taking random and epistemic uncertainty into account.
The proposed BNOR model can be implemented in evidential networks \cite{simon2009evidential}, and belief reasoning is proceeded through evoking junction tree inference algorithms \citep{simon2008bayesian,simon2009evidential}.

The remainder of this paper is organized as follows. In Section 2, the basic knowledge about
Noisy-OR gate and  Dempster--Shafer theory is briefly introduced. The BNOR model is presented in detail in Section 3.
In order to show the effectiveness
of BNOR in real practice, Section 4 discusses about how to apply BNOR on the problem of network reliability evaluation.
Conclusions are drawn in the final section.

\section{Background}

In this section some related preliminary  knowledge will be presented. The definition of Noisy-OR gate will
be described first, then some basis of belief function theory will be recalled.

\subsection{Noisy-OR model}
The Noisy-OR structure was introduced by \citet{pearl1988probabilistic} to reduce the elicitation effort in
building a Bayesian network.  The general properties of the Noisy-OR function and its generalizations were captured by
\citet{heckerman1996causal} in their definition of causal independence.
% \citet{lnor} proposed
%an extension called Leaky Noisy-OR by introducing a new parameter
%called \textquotedblleft{}leak probability\textquotedblright{}.

\begin{center} \begin{figure}[!htp] \includegraphics[width=0.45\linewidth]{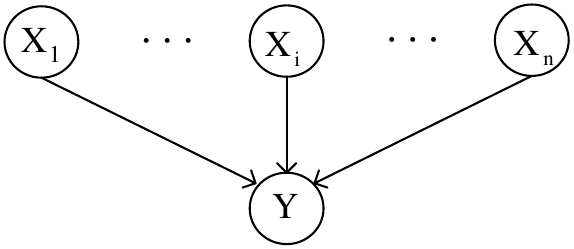}
\hfill
\includegraphics[width=0.45\linewidth]{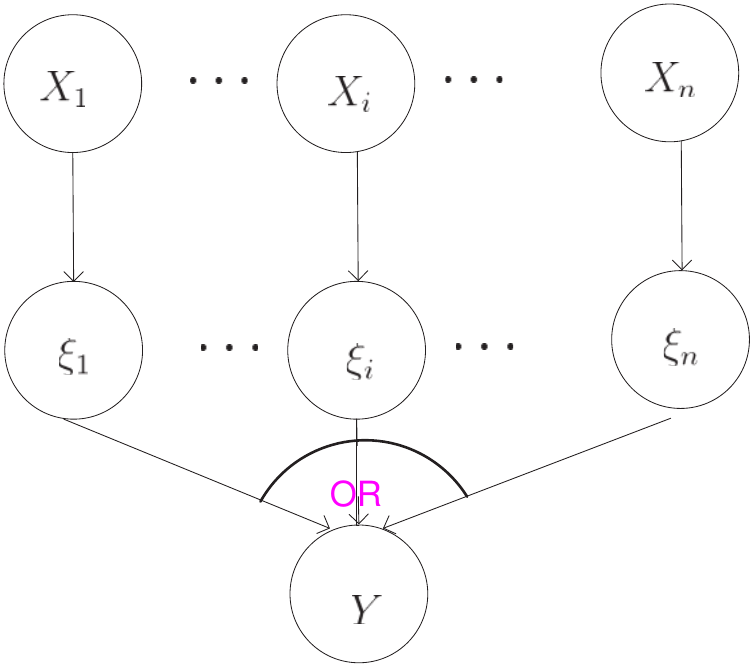}
\hfill
\parbox{.45\linewidth}{\centering\small a. causal connections} 	\hfill 	\parbox{.45\linewidth}{\centering\small b. Noisy-OR model}
\hfill
\caption{The causal connections network.}  \label{xxxy} \end{figure} \end{center}

%\begin{theorem}

Let us consider a binary variable $Y$ with $n$ binary parent variables $X_i$ (see Figure~\ref{xxxy}-a).  These variables
can be either ``True" $(T)$ or ``False" $(F)$. Each $X_{i}$ exerts its influence on $Y$
independently. To build a Bayesian network, $X$ must be associated with a probability distribution $p(Y|X_1,\cdots,X_n)$. The number of
independent parameters included in the complete specification of $p(Y|X_1,\cdots,X_n)$ is $2^n$. The Noisy-OR function is an
attractive way where we can use fewer parameters to specify $p(Y|X_1,\cdots,X_n)$. The idea
is to start with $n$ probability values $p_i$, which is the probability that $\{Y = T\}$  conditional on $\{X_i = T\}$ and $\{X_j = F\}$
for $j \neq i$, {\em i.e.,}
\begin{equation}
  p_i = p\left\{Y=T\middle| X_i = T, \left\{X_j=F\right\}^{n}_{j=1, j\neq i}\right\}.
\end{equation}
Probability $p_i$ is often called ``link probability" and illustrates the fact that the causal dependency between $X_i$ and $Y$ can be inhibited.
If  the state of variable $X_i$ is $T$, then there is chance $1-p_i$ that it is flipped to $F$; If $X_i$ is $F$, then it stays with $F$. Denote
the result of flipping (or not) $X_i$ by $\xi_{i}, i=1,2,\cdots,n$ (see Figure~\ref{xxxy}-b), then

\begin{equation}
p(Y=\alpha|X_{1,}X_{2},\cdots,X_{n})=\sum_{\alpha_{1}\vee \cdots\vee \alpha_{n}=\alpha}p(\xi_{1}=\alpha_{1}|X_{1})\cdots p(\xi_{n}=\alpha_{n}|X_{n}),\label{eq:noreq}
\end{equation}
where the values of $\alpha, \alpha_i$ are either $T$ or $F$.
%\end{theorem}
A Noisy-OR function is thus a disjunction of ``noisy" versions of $X_i$ \cite{anor2004}. Let $\bm{X}_T$ be
the set of $X_i$ whose state is ``True", and $\bm{X}_F$ be the set of $X_i$
which are ``False". The distribution of $Y$ conditional on $X_1,X_2,\cdots,X_n$ is
\begin{equation}
  p\left(Y=T\middle|X_1,X_2,\cdots,X_n\right) = 1 - \prod_{i:X_i \in \bm{X}_T}(1-p_i).
\end{equation}

We can see the number of independent parameters required  for the conditional probability function is reduced from $2^{n}$
to $2n$ \citep{introBN}. The following example shows how to create CPTs by the use of NOR gate.

%\begin{example}

\begin{center} \begin{figure}[!htp]\centering
\includegraphics[width=0.8\linewidth]{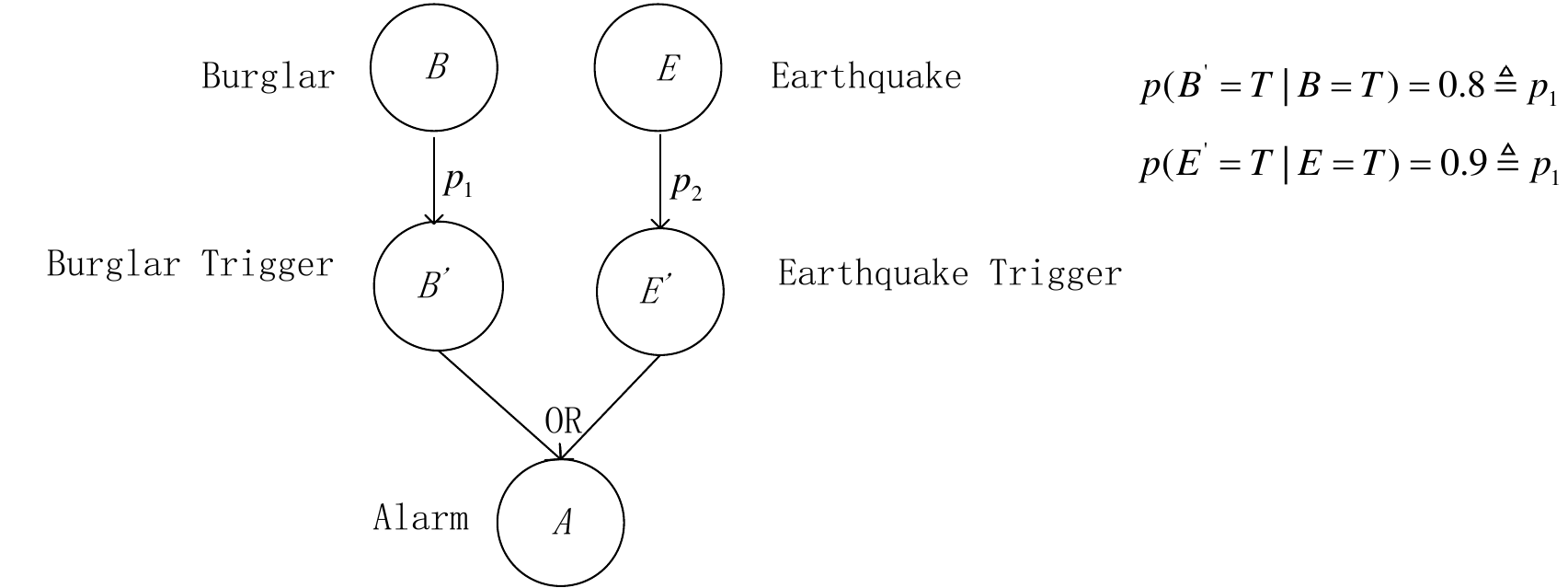}
\caption{The alarm network.}  \label{alarm} \end{figure} \end{center}

\noindent \textbf{\Exe}. Let us consider the Alarm System (see Figure~\ref{alarm}). Both a burglar
($B$) and an earthquake ($E$) can set the alarm ($A$) off but neither
always do so. The mechanism of the burglar and earthquake is different,
thus they can be regarded as independent causes. Variable $B^{'}$ (respectively, $E^{'}$)
describes the result after  flipping (or not) of $B$ (respectively, $E$). Assume all variables are binary with values $\{T,F\}$,
where $T$ represents the corresponding event  happens, while $F$ means not.

Apparently, $A$ is $F$ only if both the occurrence of burglar and earthquake
do not evoke the alarm due to inhibition. Using the Noisy-OR model, we can
get,

\begin{equation}\label{hh1}
p(A=F|B,E)=\prod_{i\in \bm{X}_{T}}(1-p_{i}),
\end{equation}
%where $X_{T}$ means the sets of variables whose state is T.
The conditional probability on $\{A=T\}$ can be obtained easily:

\begin{equation}\label{hh2}
p(A=T|B,E)=1-\prod_{i\in X_{ T}}(1-p_{i}).
\end{equation}
The following CPT can be got using Eqs.~\eqref{hh1} and \eqref{hh2}.
\begin{table}[ht] \centering \caption{The conditional probability table.}
\begin{tabular}{rrrr}
\hline
   {\it B} &    {\it E} &        $A=T$ &        $A=F$ \\
\hline
   {\it T} &    {\it T} & $1-(1-p_1)(1-p_2)$ & $(1-p_1)(1-p_2)$ \\
%\hline
   {\it T} &    {\it F} &         $p_1$ &       $1-p_1$ \\
%\hline
         F &    {\it T} &         $p_2$ &       $1-p_2$ \\
%\hline
   {\it F} &    {\it F} &          0 &          1 \\
\hline
\end{tabular}
\end{table}

%\end{example}

\subsection{Belief function theory}

To apply the theory of belief functions, we consider a set of $q$ mutually exclusive \& exhaustive elements, called the frame
of discernment, defined by
\begin{equation}
  \Theta=\{\theta_{1},\theta_{2},\cdots,\theta_{q}\}.
\end{equation}
Let $X$ be a variable taking values in $\Theta$.
The function $m:2^{\Theta}\rightarrow[0,1]$ is said to be the basic
belief assignment (bba) on $\text{2}^{\Theta}$, if it satisfies:

\begin{equation}
\sum_{A\subseteq\Theta}m(A)=1,
\end{equation}
and
\begin{equation} \label{closeworld}
  m(\emptyset) = 0.
\end{equation}
The constraint on  $\emptyset$ defined  by Eq.~\eqref{closeworld} is not mandatory. It assumes that one and only one element in $\Theta$
is true (closed-world assumption). In the case where $m(\emptyset)\neq 0$, the model accepts that none of
the elements could be true (open-world assumption) \cite{TBM}.  The closed-world assumption is accepted hereafter. Every $A\in2^{\Theta}$ such that $m(A)>0$ is called a focal element. Uncertain and imprecision knowledge about the actual value of $X$
can be represented by  a bba distributed on $2^\Theta$:
\begin{equation}
  \bm{M}_X = \left[m\left(A_1\right),m\left(A_2\right),  \cdots, m\left(A_{2^q-1}\right)\right],
\end{equation}
where $A_1, A_2, \cdots, A_{2^q-1}$ are the elements of $2^\Theta$ arranged by natural order.

%If $m$ has a single
%focal element $A$, it is said to be categorical
%and denoted as $m_{A}$. If all focal elements of $m$ are singletons,
%then $m$ is said to be Bayesian. Bayesian mass functions are equivalent
%to probability distributions.

The credibility and plausibility functions are derived from a bba $m$ as in Eqs.~\eqref{bel} and \eqref{pl}.
 %\end{definition}
\begin{equation}
Bel\text{(}A\text{)}=\sum_{B\subseteq A} m\text{(}B\text{)},~~\forall A\subseteq\Theta,
\label{bel}
\end{equation}
\begin{equation}
 Pl\text{(}A\text{)}=\sum_{B\cap A\neq\emptyset}m\text{(}B\text{)},~~\forall A\subseteq\Theta.
 \label{pl}
\end{equation}
$Bel(A)$   measures the minimal belief on $A$ justified by available information on
$B(B \subseteq A)$ , while $Pl(A)$ is the maximal belief on $A$ justified by information on $B$
which are not contradictory with $A$ ($A \cap B \neq \emptyset$). The bba can be recovered from credibility functions  through the fast M{\"o}bius transformations \citep{smets2002application}:
\begin{equation}\label{fmtbeltom}
m(A)=\sum_{B\subseteq A}(-1)^{|A-B|}Bel(B), \forall A\subseteq \Theta
\end{equation}

The relations  between $Bel$ and $Pl$ can be established as follows:
\begin{equation}\label{belandpl}
Bel(A)=1-Pl(\overline{A}) ~~,~~ Pl(A)=1-Bel(\overline{A}),
\end{equation}
where $\overline{A}$ denotes the complementary set of $A$.
$Bel(\overline{A})$ is often called the doubt in $A$. Let $Pr(A)$ denote the probability of the hypothesis $A$, it is easy to get:
\begin{equation} \label{boudper}
Bel(A) \leq Pr(A) \leq Pl(A).
\end{equation}
Probability $Pr(A)$ belongs to
the interval $[Bel(A), Pl(A)]$ but its exact value remains unknown. The bounding property \eqref{boudper} has been well
defined  in the work of Shafer \cite{ds2}.
%Many authors used it to connect the probability interval defined by $[Bel(A), Pl(A)]$, and the belief mass distribution.

%If $Pr(A)$ happens not to be precisely known, but it is known (or at least one strongly believes) that it is within $[\underline{P}(A),\overline{P}(A)]$. This leads to the use of imprecise probability \citep{walley1991statistical}, which generalizes  classical probability in the sense that uncertainty about events is quantified via intervals, instead of signal numbers \citep{coolen2004use}. $\underline{P}(A)$ is called the lower probability of the event $A$,
%$\overline{P}(A)$ the upper probability of the event $A$, and $\overline{P}(A) - \underline{P}(A)$ measures the imprecision of $A$.
%
%
%The above probability interval can be interpreted as a measure of belief. $\underline{P}(A)$ represents the
%extent to which it  is  certainly believed that $A$ is true, while $\underline{P}(\overline{A})=1-\overline{P}(A)$
%represents the extent to which it is certainly believed that $A$ is false, and the
%value $\overline{P}(A)-\underline{P}(A)$ represents the extent of the uncertainty of belief of whether
%$A$ is true or false. But what should be noticed is that although  every  belief function is a lower envelope, the  converse does not hold \citep{halpern1992two}.

%When the discernment frame is $\Theta=\{A,\overline{A}\}$,

\citet{ferson2004dependence} argued that
each Dempster-Shafer structure specifies a unique probability-box (p-box), and that  each p-box specifies an equivalent class of
Dempster-Shafer structure \cite{simon2009evidential}.
P-boxes are sometimes considered as a granular approach of imprecise probabilities \citep{walley1991statistical}, which are arbitrarily sets
of probability distributions. Probability interval $[\underline{P}(A),\overline{P}(A)]$, which is the restricted case of p-box \citep{simon2009evidential}, can also be used to describe the imprecision of a probability measure.
The relation between a probability interval and  a bba can be directly obtained \citep{simon2009evidential}:
\begin{equation}
  [\underline{P}(A),\overline{P}(A)] = [Bel(A), Pl(A)].
\end{equation}

Belief functions can be transformed into  probability distribution functions  by Smets method \cite{smets2005decision}, where each mass
of belief $m(A)$ is equally distributed among the elements of $A$. This leads to the concept of pignistic probability, $BetP$.
For all $\theta_i \in \Theta$, we have
\begin{equation}
 \label{pig}
	BetP(\theta_i)=\sum_{A \subseteq \Theta | \theta_i \in A} \frac{m(A)}{|A|(1-m(\emptyset))},
\end{equation}
where $|A|$ is the cardinality of set $A$ (number of elements of $\Theta$ in $A$).
Pignistic probabilities can help us  make a decision.

\section{Belief Noisy-OR model}
We start with the discussion of the uncertainty problem in NOR model. One of the existing approaches to express the uncertain
information in NOR is the ImNOR model proposed by \citet{fallet2012evidential}. We will analyze the drawbacks of ImNOR
and present a new NOR gate using the theory of belief functions.
\subsection{The uncertainty problem in NOR structure}

%Let us consider the alarm network mentioned before. Usually it is
%difficult to determine whether an earthquake has happened, especially
%when the magnitude is relatively small and the hypo-center is deep.

From an industrial point of view, it is
classically accepted that observations made on the system
are partially realized \cite{utkin2007imprecise}. For example, it is
difficult to determine whether an earthquake has happened, especially
when the magnitude is  small and the hypo-center is deep. In such a case, there is  some uncertainty on the state of boolean
parent variables and it is intuitive for experts to give a positive belief on the ignorant modality $\left\{T,F \right\}$. \citet{simon2009evidential} have investigated a solution based
on evidential network  and the theory of belief functions to
take into account the uncertainty on the state of binary parent
variables in AND/OR gates.
\citet{simon2008bayesian} combined belief function theory with Bayesian reasoning to deal with
this type of epistemic uncertainty.
Based on Simon and Weber's modelling formalization, \citet{fallet2012evidential}
proposed imprecise extensions of the Noisy-OR (ImNOR) structure to
deal with the uncertainty on the state of variables and link probabilities.

ImNOR describes the uncertainty on variable state and link probabilities separately
by calculating the lower bounds of the conditional probability
$P\left(X\middle|\text{Pa}(X)\right)$, where $\text{Pa}(X)$ denotes the parent nodes
of $X$. Consider the causal network shown in Figure \ref{xxxy}-a. As before, the discernment frame of each variable
is $\left\{T,F \right\}$, and  each $X_i$ is interpreted as an independent  ``cause" of $Y$.
%State $T$ means that the causes that the variable represents occurs while $F$ means not.
We can express our epistemic uncertainty on variables' state by assigning
the basic belief to ignorant modality $\left\{T,F\right\}$.  This modality
indicates that the variable  is exclusively in $\{T\}$ or $\{F\}$ state
without distinguishing exactly in which state it is \cite{fallet2012evidential}.
%The fail of either $X_{i}$ $(X_{i}= F)$ could cause the system's fail ($Y= F$).
Different from Noisy-OR model, in ImNOR, each active $X_i$  can evoke $Y$ with
unknown probability $p_i \in [p_{iL},p_{iU}]$, where $p_{iU}-p_{iL}$ measures
the degree of uncertainty on our knowledge of inhibition. Fallet
provided us the formulas (see Eqs.~\eqref{eq:simoneq1}--\eqref{eq:simoneq3}) to calculate the conditional belief mass
functions\footnote{As the belief functions are defined on the power set of discernment frame, we use $\{T\}$  ($\{F\}$) instead of
$T$ ($F$) to denote variable state here.}:

\begin{equation}
m(Y= \{T\}|X_{1},X_{2},\cdots,X_{n})=1-\prod_{\{i:X_{i}= \{T\}\}}(1-p_{iL}),\label{eq:simoneq1}
\end{equation}

\begin{equation}
m(Y= \{F\}|X_{1},X_{2},\cdots,X_{n})=\prod_{\{i:X_{i}= \{T\}\}}(1-p_{iU})\prod_{\{i:X_{i}=\text{\ensuremath{\left\{   T,F  \right\} } }\}}(1-p_{iU}),\label{eq:simoneq2}
\end{equation}

\begin{equation}
m(Y=\text{\ensuremath{\left\{ T,F  \right\} } }|X_{1},X_{2},\cdots,X_{n})=\prod_{\{i:X_{i}= \{T\}\}}(1-p_{iL})-\prod_{\{i:X_{i}= \{T\}\}}(1-p_{iU})
\prod_{\{i:X_{i}=\text{\ensuremath{\left\{T,F \right\} } }\}}(1-p_{iU}).\label{eq:simoneq3}
\end{equation}
%\begin{center} \begin{figure}[!htp] \centering
%\includegraphics[width=0.6\linewidth]{ccc_s.pdf} \caption{The imprecise Noisy-OR model}  \label{cccs} \end{figure} \end{center}
From Eq.~\eqref{eq:simoneq1}, we can get:
\begin{equation}
  m\left(Y= \{T\} \middle|X_{1} \cdots X_{k}=\left\{T,F \right\} \cdots X_{n}\right) =  1-\prod_{\{i:X_i= \{T\}, i \in \{1,2,\cdots,k-1, k+1, \cdots, n\}\}}(1-p_{iL}),
\end{equation}
and
\begin{equation}
  m\left(Y= \{T\} \middle|X_{1} \cdots X_{k}=\left\{F \right\} \cdots X_{n}\right) =  1-\prod_{\{i:X_i= \{T\}, i \in \{1,2,\cdots,k-1, k+1, \cdots, n\}\}}(1-p_{iL}).
\end{equation}
It can be seen that the belief on the proposition that ``variable  $Y$ is $\{T\}$" does
 not change when some prior precise information about the parent variables becomes available (The state of $X_k$ changes from
 $\{T,F\}$ to $\{T\}$):
\begin{equation}
m(Y= \{T\}|X_{1}, \cdots, X_{i}=\{ T,F \}, \cdots X_{n})= m(Y= \{T\}|X_{1}, \cdots, X_{i}=\{ F\}, \cdots, X_{n}).\label{eq:unreasonable1}
\end{equation}
%If we have some new information indicating that cause $X_k$ does not happen, it is intuitive that we have more belief
%Similarity we have
%\begin{equation}
%m(Y= F|X_{1} \cdots X_{i}=\{ T,F \} \cdots X_{n})= m(Y= F|X_{1} \cdots X_{i}=\{ T\} \cdots X_{n})\label{eq:unreasonable2}
%\end{equation}

 \iffalse
 we could not use Eq.~\eqref{eq:simoneq1} since there is no $X_{i}$
satisfies $X_{i}=T$.

Also we could not regard that $\prod_{X_{i}= T}(1-p_{iL})=0$,
because if so, the following belief distributions could be derived:

\begin{equation}
P(Y= T|X_{i}= F,i=1,2,\cdots,n-1,X_{n}=\left\{  T,F \right\} )=1,
\end{equation}

\begin{equation}
P(Y= F|X_{i}= F,i=1,2,\cdots,n-1,X_{n}=\left\{  T,F \right\} )=1-P_{nU}>0,
\end{equation}
Obviously,
\[
P(Y= T|X_{i}= F,i=1,2,\cdots,n-1,X_{n}=\left\{  T,F \right\} )+P(Y= F|X_{i}= F,i=1,2,\cdots,n-1,X_{n}=\left\{  T,F \right\} )>1.
\]

Besides, $P(Y= T|X_{i}= F,i=1,2,\cdots,n-1,X_{n}=\left\{  T,F \right\} )=1$
is unreasonable, as we could not say that the system is at work with
a belief 100\% when the only information is that all the components
fail except one we do not have any knowledge of its state. Thus this
basic belief assignment approach by ImNOR could not be used in these
cases.
\fi
Besides, there is another defect for the above method when calculating the
belief mass on the conditional events where there is no working components $(X_{i}\neq T,i=1,2,\cdots,n)$. For example, when we want
to know
\[
m(Y|X_{i}= \{F\},i=1,2,\cdots,n-1,X_{n}=\left\{  T,F \right\} ),
\]
the following conditional belief mass functions can be got by Eq.~\eqref{eq:simoneq2}:
%\iffalse
%\begin{equation}
%m(Y= \{T\}|X_{i}= \{F\},i=1,2,\cdots,n-1,X_{n}=\left\{T,F \right\} )=0,
%\end{equation}
\begin{equation}
m(Y= \{F\}|X_{i}= \{F\},i=1,2,\cdots,n-1,X_{n}=\left\{T,F \right\} )=1-P_{nU},
\end{equation}
\begin{equation}
m(Y= \{F\}|X_{i}=\text{\{T,F\}},i=1,2,\cdots,n-1,X_{n}=\left\{T,F \right\} )=P_{nU}.
\end{equation}
%\fi
As can be seen, the lower bound of probability $p_n$, $p_{nL}$, has no effect on the final results. That is to say, the conditional belief mass assignment remains unchanged once the upper bound of $p_n$ is fixed no matter how long the uncertain interval is.  This  is against  our common sense. Since the length of the interval measures the degree of uncertainty on the available information, the longer the interval is, more mass value should be given to the ignorant state $\{ T,F\}$.

\subsection{Belief Noisy-OR structure}
We introduce here the Belief Noisy-OR
structure to express the epistemic
uncertainty on the state of the variables and link probabilities at the same time.
Consider the causal network where $X_i, i=1,2,\cdots,n$ are the parents of $Y$ (Figure \ref{xxxy}-a).

%In the framework of belief functions, we can express our knowledge on the
%state by defining belief functions on the power set $2^\Theta=\{ T, F,\{ T,F \}\}$. The uncertainty on the parameters is first represented by lower and upper probabilities, and then the basic belief assignment is defined accordingly.

%The probability $p_i$ corresponds to the probability that $Y$ is true if $X_i$ is true, {\em i.e.,}
%\begin{equation}
%  p_i = P\left(Y = \{T\} \mid X_i = \{T\}, X_{j, j\neq i} = \{F\}\right).
%\end{equation}
Let us first discuss the uncertainty on link probability $p_i$. This parametric uncertainty can be modeled by an interval
$[p_{iL},p_{iU}]$, with $0 \leq p_{iL}\leq p_i \leq p_{iU} \leq 1$. Thus the inhibition probability interval
of $X_{i}$ is $[1-p_{iU},1-p_{iL}]$.
%Let $I_{i}$ be a variable indicating whether
%the inhibition occurs. $\mbox{\ensuremath{I_{i}= T}}$ denotes
%there is no inhibition and $Y$ will surely be in state $T$ if $X_{i}= T$.
%Nevertheless, $Y$ may be $F$ even $X_{i}$ is in the normal state ($X_{i}= T$) when
%$I_{i}= F$. $I_{i}$ is considered in the state $\left\{  T,F \right\} $
%%\textcolor{red}{when we could not distinguish its state from $ T$ to F}
%when it is exclusively in one of the previous
%conditions (T and F) without distinguishing exactly which. For
%convenience, $n$ auxiliary variables, $X_{i}^{'},i=1,2,\cdots,n$
%are introduced to represent the state of $X_{i}$ after the inhibition.
\iffalse
\begin{equation}
\underline{P}(A)=Bel(A) ~~ and~~ \overline{P}(A)=Pl(A).
\label{interval2belief}
\end{equation}
Using Eq.~\eqref{interval2belief}
\fi
In order to use  evidential reasoning, the probability intervals should be transformed to belief function structures.  For
convenience, $n$ auxiliary variables, $X_{i}^{'},i=1,2,\cdots,n$
are introduced to represent the result of flipping or not  $X_{i}$. From the boundary
property of belief functions  (Eq.~\eqref{boudper}), we can get
\begin{equation}
Bel(X_{i}^{'}= \{T\}|X_{i}= \{T\})=p_{iL},
\end{equation}

\begin{equation}
Pl(X_{i}^{'}= \{T\}|X_{i}= \{T\})=p_{iU},
\end{equation}

%
%In fact  $p_{iL}$ represents the
%extent to which we believe the event $X_i^{'}=\mathrm{T}$ on the condition that $X_i=\mathrm{T}$, while $1-p_{iU}$ represents the extent
%to which we believe the event $X_i^{'}=\mathrm{F}$ conditioned that $X_i=\mathrm{T}$, and the value $p_{iU}- p_{iL}$ represents the extent of the uncertainty of belief about whether $X_i^{'}$ is true or false.

The  associated belief mass distribution can be easily
determined by Eq.~\eqref{fmtbeltom}. The corresponding  conditional bba in $2^\Theta$ can be defined as:
 %(we still use the symbol  $p$ instead of $m$ when no ambiguity is possible):

\begin{equation}
m(X_{i}^{'}= \{T\}|X_{i}= \{T\})= Bel(X_{i}^{'}= \{T\}|X_{i}= \{T\}) = p_{iL},\label{eq:ccceq1}
\end{equation}

\begin{align}
m(X_{i}^{'}= \{F\}|X_{i}= \{T\}) &= Bel(X_{i}^{'}= \{F\}|X_{i}= \{T\})\\ &= 1 - Pl(X_{i}^{'}= \{T\}|X_{i}= \{T\}) \label{equa}\\&= 1-p_{iU},
\end{align}
and
\begin{equation}
m(X_{i}^{'}=\text{\ensuremath{\left\{T,F \right\} } }|X_{i}= \{T\})=p_{iU}-p_{iL}, \label{eq:ccceq2}
\end{equation}
where Eq.~\eqref{equa} is obtained by the relation between $Bel$ and $Pl$. Eqs.~\eqref{eq:ccceq1}--\eqref{eq:ccceq2} express the uncertain knowledge of link probabilities and variable states together  in the form of belief mass distributions.
%$p_{iU}-p_{iL}$ measures our uncertainty of the interval of the $p_i$, in fact, it is our belief given to the state of $I_i=\{ T,F \}$.

When $X_{i}= \{F\}$, $Y$ is sure to stay in state $\{F\}$ .
\iffalse
\begin{equation}
Bel(X_{i}^{'}= T|X_{i}= F)=Pl(X_{i}^{'}= T|X_{i}= F)=0.
\end{equation}
\begin{equation}
Bel(X_{i}^{'}= F|X_{i}= F)=Pl(X_{i}^{'}= F|X_{i}= F)=1.
\end{equation}
\fi
Thus the bba conditioned $X_i= \{F\}$ is easy to determine:
\begin{equation}
m(X_{i}^{'}= \{T\}|X_{i}= \{F\})=0,
\end{equation}
\begin{equation}
m(X_{i}^{'}= \{F\}|X_{i}= \{F\})=1,
\end{equation}
\begin{equation}
m(X_{i}^{'}=\text{\{T,F\}}|X_{i}= \{F\})=0.
\end{equation}
However, the bba on the condition $X_i=\text{\{T,F\}}$ is more complicated. The following equations with unknown parameters $\alpha,\beta,\gamma ~(\alpha+\beta+\gamma=1)$ are first given, and then the methods for designing the three parameters will be discussed later.

\begin{equation}
m(X_{i}^{'}=\text{\ensuremath{\{T\}}}|X_{i}=\text{\ensuremath{\left\{T,F \right\}}})=\alpha,\label{eq:difftosimon}
\end{equation}

\begin{equation}
m(X_{i}^{'}=\text{\ensuremath{\left\{F \right\}}}|X_{i}=\text{\ensuremath{\left\{T,F \right\}}})=\beta,\label{eq:difftosimon2}
\end{equation}

\begin{equation}
m(X_{i}^{'}=\text{\ensuremath{\left\{T,F\right\}}}|X_{i}=\text{\ensuremath{\left\{T,F \right\}}})=\gamma.\label{eq:difftosimon3}
\end{equation}
Eqs.~\eqref{eq:difftosimon}--\eqref{eq:difftosimon3} show that in  BNOR,  the belief on the uncertain state $X_i=\{ T,F \}$ may be flipped into
all the possible states by different ratios. Parameters $\alpha,\beta,\gamma$ are adjustable.
%It is worthy of notice that there are two kinds of uncertainty information in the network here, one is on the inhibition parameters,
%represented in the form of imprecise intervals and expressed in our method in the belief mass assigned to  $I_i=\{ T,F \}$. Another is on  on  the  state  of parent   variables $X_i$, represented in the vague  state $X_i=\{ T,F \}$.
%It is important to notice that BNOR can deal with two kinds of uncertainty. One is on the link probabilities, represented in the form of
%probability intervals. The other is on the state of  variables, represented in the vague state $X_i=\{ T,F \}$.

Generally, the values of $\alpha,\beta,\gamma$  can be given by the proportions of belief mass on $X_i=\{ T,F \}$ which may be transferred to $X_i= \{T\}$ (noted by $\lambda_1$, $0 \leq \lambda_1 \leq 1$),  $X_i= \{F\}$ (noted by $\lambda_2$, $0 \leq \lambda_2 \leq 1$) and $X_i=\{ T,F \}$ (noted by $\lambda_3$, $0 \leq \lambda_3 \leq 1$) ($\lambda_1+\lambda_2+\lambda_3=1$) respectively:
\begin{equation}
\alpha=\lambda_1 m(X_{i}^{'}=\text{\ensuremath{\{T\}}}|X_{i}=\text{\ensuremath{\{T\}}})=\lambda_1 p_{iL},\label{eq:smetspig010}
\end{equation}

\begin{equation}
\beta=\lambda_1 m(X_{i}^{'}=\text{\ensuremath{\{F\}}}|X_{i}=\text{\ensuremath{\{T\}}})+\lambda_2= \lambda_1 (1-p_{iU})+\lambda_2,\label{eq:smetspig020}
\end{equation}

\begin{equation}
\gamma=\lambda_1 m(X_{i}^{'}=\text{\ensuremath{\left\{T,F \right\}}}|X_{i}=\text{\ensuremath{\{T\}}})+\lambda_3= \lambda_1(p_{iU}-p_{iL})+\lambda_3.\label{eq:smetspig030}
\end{equation}
Parameter $\lambda_3$ in Eq.~\eqref{eq:smetspig030} indicates the uncertainty on the state of $X_i$, and it should be in direct
proportion to $m(X_i=\{ T,F \}) \triangleq \eta$. It is easy to know that if $ \eta \neq 0$, $\lambda_3 \neq 0$. For simplicity, let $\lambda_3=\eta$ and $\lambda_1=\lambda$, $\lambda_2=1-\lambda-\eta$, then,

\begin{equation}
\alpha=\lambda m(X_{i}^{'}=\text{\ensuremath{ \{T\} }}|X_{i}=\text{\ensuremath{ \{T\}}})=\lambda p_{iL},\label{eq:smetspig10}
\end{equation}

\begin{equation}
\beta=\lambda m(X_{i}^{'}=\text{\ensuremath{  \{F\}  }}|X_{i}=\text{\ensuremath{ \{T\} }})+(1-\lambda-\eta)= \lambda (1-p_{iU})+(1-\lambda-\eta),\label{eq:smetspig20}
\end{equation}

\begin{equation}
\gamma=\lambda m(X_{i}^{'}=\text{\ensuremath{\left\{   T,F  \right\}}}|X_{i}=\text{\ensuremath{  \{T\}}})+\eta= \lambda(p_{iU}-p_{iL})+\eta.\label{eq:smetspig30}
\end{equation}
This is similar to the optimistic coefficient method in the decision theory. So we call this general approach Optimistic coefficient-BNOR (OCBNOR), where $\lambda~ (0 \leq \lambda \leq 1)$ is the optimistic coefficient.

Note that Eqs.~\eqref{eq:smetspig10}--\eqref{eq:smetspig30} propagate the uncertainty on variable state and link probabilities simultaneously.  The
ignorant  modality $\Theta = \{T,F\}$ represents the uncertainty on the state and the belief mass assignment  $m(X_i^{'}|X_i)$ deals with the uncertain  information of link probabilities.

Different $\lambda$ values can be set to obtain results under various requirements. For instance, if we want to make an optimistic decision, the belief to  $X_i=\{ T,F \}$ could  transferred to  $X_i^{'}= \{T\}$ as most as possible. Let $\lambda=1$, then
\begin{equation}\alpha=p_{iL},\beta=1-p_{iU}-\eta,\gamma=p_{iU}-p_{iL}+\eta.\end{equation}
This structure is called Optimistic Belief Noisy-OR (OBNOR).
By contrary, when a pessimistic decision is required, all the belief on $X_i=\{ T,F \}$ could transferred to the child state $X_i^{'}= \{F\}$,  thus
 \begin{equation}\alpha=0,\beta=1-\eta,\gamma=\eta, \end{equation}
we call this model Pessimistic Belief Noisy-OR (PBNOR).

The decision-makers can make a compromise between optimism and pessimism. According to the idea of pignistic probability transformation \cite{smets2005decision}, the belief on the subsets of the discernment framework should be given to the single elements equally. Then we can get:

\begin{equation}
\alpha=\frac{1}{2} m(X_{i}^{'}=\text{\ensuremath{   \{T\} }}|X_{i}=\text{\ensuremath{ \{ T\}}})=\frac{1}{2}p_{iL},\label{eq:smetspig1}
\end{equation}

\begin{equation}
\beta=\frac{1}{2}m(X_{i}^{'}=\text{\ensuremath{ \{ F\}  }}|X_{i}=\text{\ensuremath{\{ T\} }})+\frac{1}{2}-\eta=\frac{1}{2}(1-p_{iU})+\frac{1}{2}-\eta,\label{eq:smetspig2}
\end{equation}

\begin{equation}
\gamma=\frac{1}{2}m(X_{i}^{'}=\text{\ensuremath{\left\{   T,F  \right\} }}|X_{i}=\text{\ensuremath{ \{ T\}}})+\eta= \frac{1}{2}(p_{iU}-p_{iL})+\eta.\label{eq:smetspig3}
\end{equation}
BNOR with the above $\alpha,\beta,\gamma$ is called Temperate Belief Noisy-OR (TBNOR) model. It is easy to see that OBNOR, PBNOR and TBNOR are special cases of OCBNOR.

Least Committed Belief Noisy-OR (LC-BNOR), just as the name implies, suggests us that the least
committed belief mass function should be selected holding the view that one should never give more belief than justified. It satisfies a form of skepticism, of noncommitment, of
conservatism in the allocation of the beliefs \citep{smets2005belief}.  At this time,
\begin{equation}\alpha=0,\beta=0,\gamma=1.\end{equation}

\iffalse
This seems more
reasonable than Fallet's ImNOR, where the uncertainty of $X_{i}$
may transfer to the determinate states of $X_{i}^{'}$. \fi

\subsection{From Bayesian Networks to Evidential Networks}
In order to apply BNOR structure,
it is important to find a relevant model to encode and to
propagate the causal relations included in BNOR. Here the evidential network model proposed by \citet{simon2009evidential} is taken as a solution to implement BNOR.
Similar to BN, evidential networks allow dealing
with a lot of variables and modeling the dependencies  between  variables. From BNOR, the conditional mass distribution
$$ m(Y=\{T\}|X_i),~~ m(Y=\{F\}|X_i), ~~m(Y=\{T,F\}|X_i)$$
can be established, which plays a similar role as CPTs in Bayesian networks.
%Then the  junction tree inference algorithms can be evoked
%to obtain the mass assigned to the state of $Y$ conditional on some prior knowledge on states of $X_i$.
If the prior belief mass values of the parent nodes $X_1,\cdots,X_n$ are given, then the junction tree inference algorithm
can be evoked to calculate the marginal mass distribution of the child node $Y$. Once the bba of $Y$ is got, the belief and plausibility functions of $Y$ can be obtained accordingly.

\subsection{The Pignistic probability and decision making}
The transferable belief model (TBM) \citep{smets2005decision} is an interpretation of the Dempster-Shafer theory of evidence, where beliefs can be held at two levels --- credal level and pignistic level. When an agent has to select an optimal action among an exhaustive set of actions, rationality principles lead to the use of a probability measure. Therefore, when a decision has to be made, the bba
obtained by BNOR model must be transformed into a
probability measure \citep{mercier2005fusion}.  One of the most commonly used transformation approaches is Smets method shown in Eq.~\eqref{pig}. %. In this algorithm, the probability that each element occurs is considered to be equal. As a result, the bba of each set is assigned to the element it contained equally \citep{smets2005decision}:
%\begin{equation}
% BetP(\theta_i)=\sum_{\theta_i  \in A \subseteq \Theta } \frac{m(A)}{|A|(1-m(\emptyset))},
%\end{equation}
%where $|A|$ is the number of elements of $\Theta$ in $A$.
In our cases, $\Theta=\{ T,F \}$. If the following bba is got,
\begin{equation}
m(X= \{T\})=m_1, m(X= \{F\})=m_2,m(X=\text{\{T,F\}})=m_3,
\end{equation}
we can get the following pignistic probability:
\begin{equation}
 BetP(X= T)=m_1+\frac{m_3}{2},~~
 BetP(X= F)=m_2+\frac{m_3}{2}.
\label{pignistic}
\end{equation}
If the conditional belief mass functions is obtained,
\begin{equation}
m(Y= \{T\}|X)=m_1^X, m(Y= \{F\}|X)=m_2^X,m(Y=\text{\{T,F\}}|X)=m_3^X,
\end{equation}
the  conditional pignistic probability can be got as follows:
\begin{equation}
 BetP(Y= T|X)=m_1^X+\frac{m_3^X}{2},~~
 BetP(Y= F|X)=m_2^X+\frac{m_3^X}{2}.
 \label{pignistic1}
\end{equation}

%\begin{example}

\begin{center}
 \begin{figure}[!htp] 	\includegraphics[width=0.45\linewidth]{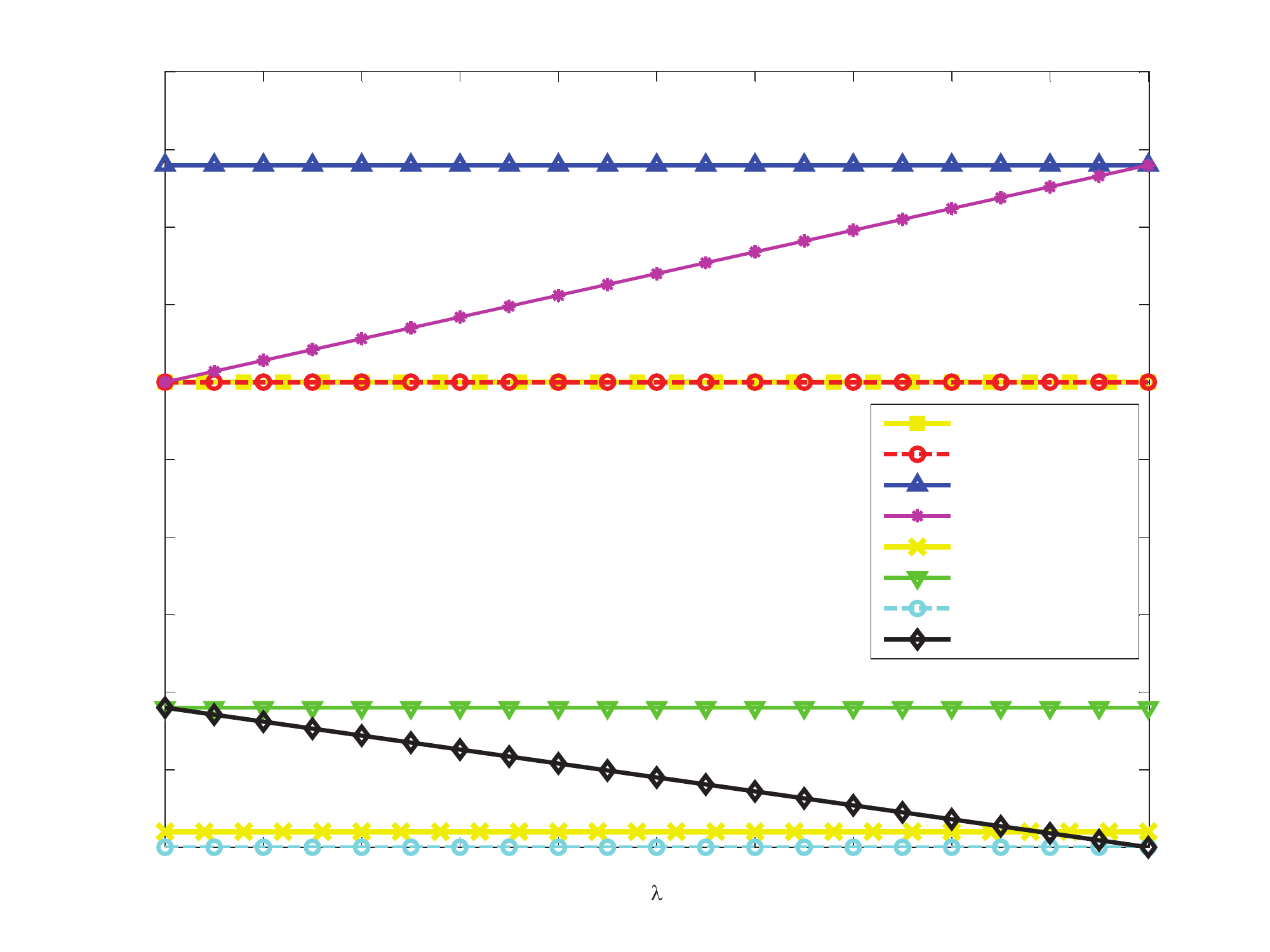} 	\hfill 	\includegraphics[width=.45\linewidth]{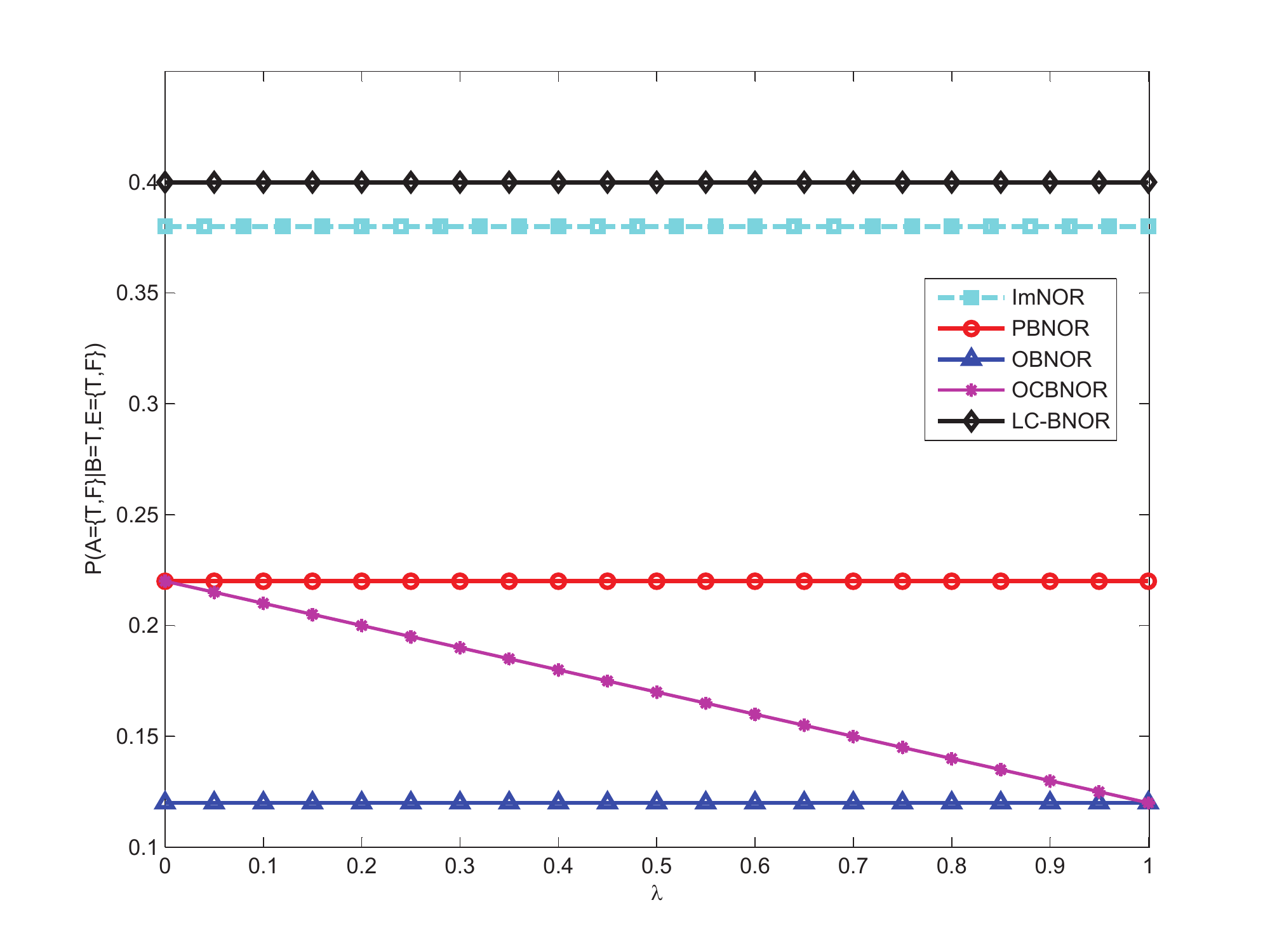} 	\hfill
 \parbox{.45\linewidth}{\centering\small a. $m(A=\{ T( F)\}|B= T, E=\{ T,F \})$}
	\hfill
	\parbox{.45\linewidth}{\centering\small b. $m(A=\{ T,F \}|B= T, E=\{ T,F \})$}
	\hfill
 \caption{The different BNOR structures for $m(A|B= T,E=\text{\{T,F\}})$.}
 \label{example0606} \end{figure} \end{center}

\noindent \textbf{\Exe}. Here we use the example of Alarm System again to illustrate the behavior of different BNOR models
and ImNOR.
The intervals of the link probabilities of  burglar and earthquake are $p_{1} \in (0.6,0.8)$ and $p_{2} \in (0.7,0.9)$
respectively. The prior distribution of $B$ and $E$ are given:
$$m(B= \{T\})=0.4, m(B= \{F\})=0.6, m(B=\text{\ensuremath{\left\{   T,F  \right\} }})=0,$$
$$m(E= \{T\})=0.3, m(E= \{F\})=0.6, m(E=\text{\ensuremath{\left\{   T,F  \right\} }})=0.1.$$
%\end{example}
The corresponding BNOR model is shown in Figure~\ref{alarm_belief}.
The conditional mass function $m(A|B,E)$ based on different BNOR models and ImOR are displayed
in Tables~\ref{bnorcpdf}--\ref{bnorcpdf-b-tf}. Figure~\ref{example0606} illustrates the value of
$m(A|B= \{T\},E=\text{\{T,F\}})$ by different schemes.
It can be seen that, ImNOR provides a pessimistic  decision for $m(A= \{T\}|\cdot)$, but an optimistic one for $m(A= \{F\}|\cdot)$. This is
counter-intuitive as ImNOR holds opposite  attitudes towards one event.
\begin{center} \begin{figure}[!htp] \centering
\includegraphics[width=0.6\linewidth]{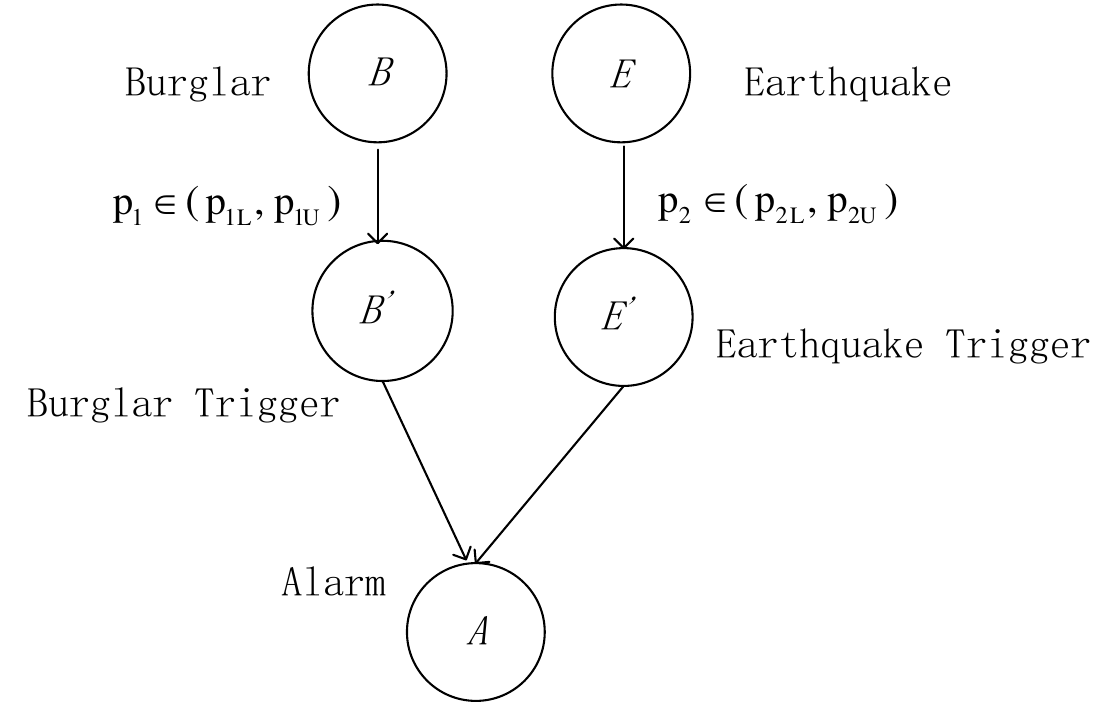}
\caption{The Belief Noisy-OR structure for Alarm System.}  \label{alarm_belief} \end{figure} \end{center}
From Table ~\ref{bnorcpdf} we can see, by the use of ImNOR,

\begin{equation}
m(A= \{F\}|B= \{T\},E= \{T\})=m(A= \{F\}|B= \{T\},E=\{ T,F \}),
\label{imnor_not}
\end{equation}
but by BNOR,
\begin{equation}
m(A= \{F\}|B= \{T\},E= \{T\}) < m(A= \{F\}|B= \{T\},E=\{ T,F \}).
\label{bnor_not}
\end{equation}
Eq.~\eqref{bnor_not} fits  with our common sense. As soon as  we know the earthquake has   happened for sure,
the less belief should be holding for the proposition the alarm  has not gone.

\begin{center}
 \begin{figure}[!htp] 	
 \includegraphics[width=0.45\linewidth]{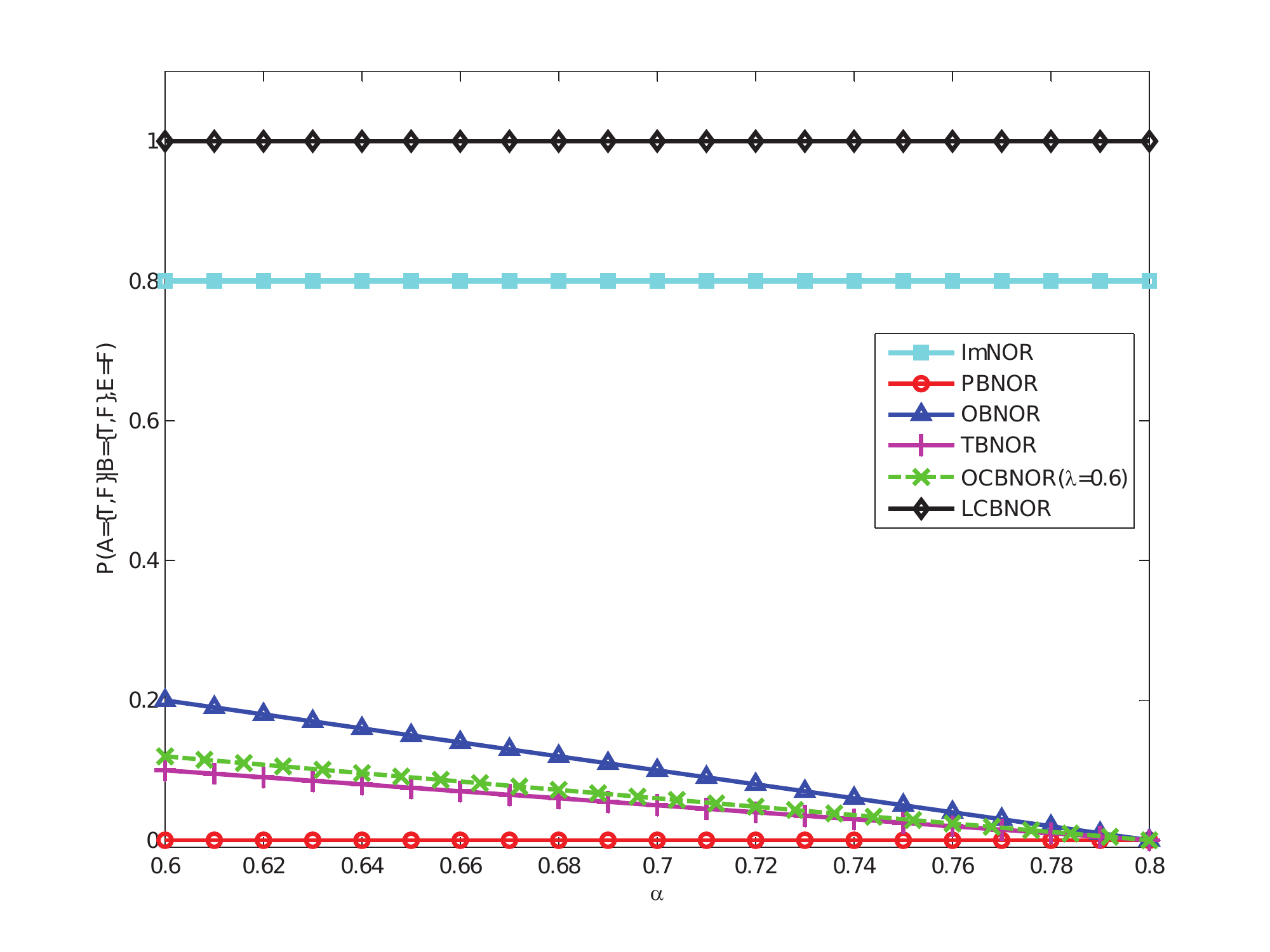} 	\hfill 	\includegraphics[width=.45\linewidth]{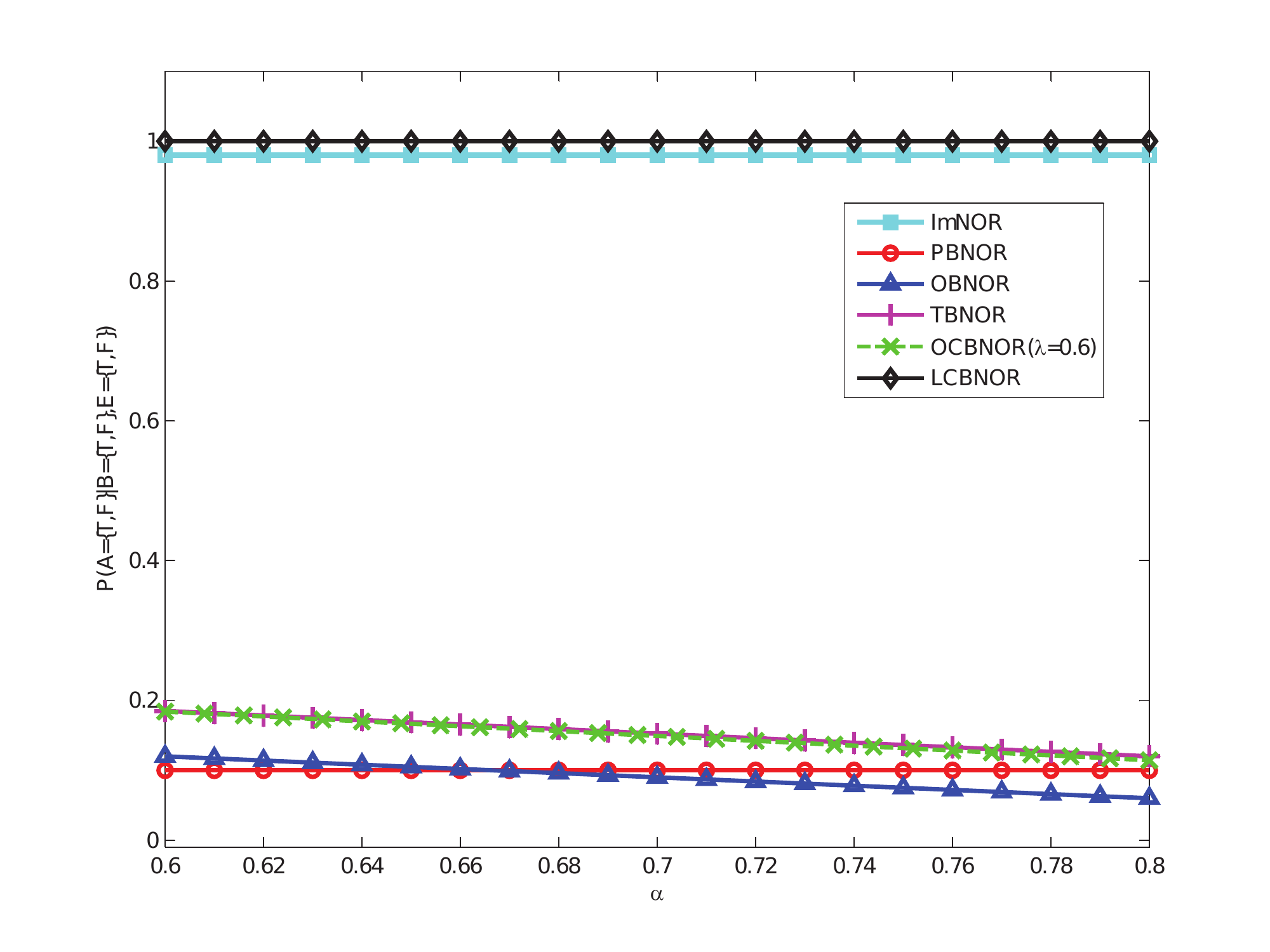} 	\hfill
 \parbox{.45\linewidth}{\centering\small a. $m(A=\{ T,F \}|B=\{ T,F \},E= F)$}
	\hfill
	\parbox{.45\linewidth}{\centering\small b. $m(A=\{ T,F \}|B=\{ T,F \},E=\{  T,F \})$}
	\hfill
 \caption{The uncertainty of the state of the alarm.}
 \label{intervallength} \end{figure} \end{center}

Let the link probability  of earthquake be $p_2 \in (0.7,0.9)$, while the link probability  of  burglar
is set to be $p_1 \in (\alpha,0.8)$. When $\alpha \rightarrow 0.8$, the
uncertainty on $p_1$ becomes less. Consequently the uncertainty  on the state of Alarm should also decrease. However, as shown in Figure~\ref{intervallength}, the values of $m(A=\{ T,F \}|B=\{ T,F \},E= \{F\})$ and $m(A=\{ T,F \}|B=\{ T,F \},E=\{T,F \})$ by ImNOR remain unchanged
although the information for $p_1$ becomes more precise. The corresponding results using BNOR show that the  belief mass assigned to the uncertainty state of $A$ decreases with the  increasing precision of the information on $p_1$. Specially, if there is no epistemic uncertainty on $p_2$, and the uncertainty on $p_1$ declines to 0 ($\alpha=0.8$),  the belief mass given to $A=\{ T,F \}$ also becomes zero (see Figure~\ref{intervallength}-a).

\begin{table}[htp] \centering \caption{The conditional mass distribution for $P(A|B= T,E)$ by ImNOR and different BNORs.} \begin{tabular}{rrrrrrrrr} \hline
         $A$ &  \multicolumn{ 3}{c}{($B$,$E$) of ImNOR} &            & \multicolumn{ 3}{c}{($B$,$E$) of LC-BNOR} \\

           &      ($T$,$T$) &      ($T$,$F$) &  ($T$,$\{T,F\}$) &            &      ($T$,$T$) &      ($T$,$F$) &  ($T$,$\{T,F\}$) \\
           \cline{2-4}   \cline{6-8}

         $T$ &    0.8800  &    0.6000  &    0.6000  &            &    0.8800  &    0.6000  &    0.6000  \\

         $F$ &    0.0200  &    0.2000  &    0.0200  &            &    0.0200  &    0.2000  &    0.0000  \\

   $\{T,F\}$ &    0.0200  &    0.2000  &    0.3800  &            &    0.0200  &    0.2000  &    0.4000  \\

\hline
           &  \multicolumn{ 3}{c}{($B$,$E$) of PBNOR} &            &  \multicolumn{ 3}{c}{($B$,$E$) of OBNOR} \\

           &      ($T$,$T$) &      ($T$,$F$) &  ($T$,$\{T,F\}$) &            &      ($T$,$T$) &      ($T$,$F$) &  ($T$,$\{T,F\}$) \\
 \cline{2-4}   \cline{6-8}

         $T$ &    0.8800  &    0.6000  &    0.6000  &            &    0.8800  &    0.6000  &    0.8800  \\

         $F$ &    0.0200  &    0.2000  &    0.1800  &            &    0.0200  &    0.2000  &    0.0000  \\

   $\{T,F\}$ &    0.0200  &    0.2000  &    0.2200  &            &    0.0200  &    0.2000  &    0.1200  \\

\hline
           &  \multicolumn{ 3}{c}{($B$,$E$) of TBNOR} &            & \multicolumn{ 3}{c}{($B$,$E$) of OCBNOR($\lambda$=0.6)} \\

       &      ($T$,$T$) &      ($T$,$F$) &  ($T$,$\{T,F\}$) &            &      ($T$,$T$) &      ($T$,$F$) &  ($T$,$\{T,F\}$) \\
 \cline{2-4}   \cline{6-8}

         $T$ &    0.8800  &    0.6000  &    0.7400  &            &    0.8800  &    0.6000  &    0.7680  \\

         $F$ &    0.0200  &    0.2000  &    0.0900  &            &    0.0200  &    0.2000  &    0.0720  \\

   $\{T,F\}$ &    0.0200  &    0.2000  &    0.1700  &            &    0.0200  &    0.2000  &    0.1600  \\

\hline
\end{tabular} \label{bnorcpdf} \end{table}
\iffalse
\begin{center} \begin{figure}[!htp] \includegraphics[width=1\linewidth]{EV1.pdf} \label{alarm_belief} \caption{The Belief-Noisy-OR structure of the alarm network} \end{figure} \end{center}
\fi

\begin{table}[htp] \centering \caption{The conditional mass distribution for $P(A|B= F,E)$ by ImNOR and different BNORs.} \begin{tabular}{rrrrrrrrr} \hline
         A &  \multicolumn{ 3}{c}{(B,E) of ImNOR} &            & \multicolumn{ 3}{c}{(B,E) of LC-BNOR} \\

           &      ($F$,$T$) &      ($F$,$F$) &  ($F$,\{$T$,$F$\}) &            &      ($F$,$T$) &      ($F$,$F$) &  ($F$,\{$T$,$F$\}) \\
           \cline{2-4}   \cline{6-8}

         $T$ &    0.7000  &    0.0000  &    0.0000  &            &    0.7000  &    0.0000  &    0.0000  \\

         $F$ &    0.1000  &    1.0000  &    0.1000  &            &    0.1000  &    1.0000  &    0.0000  \\

   \{$T$,$F$\} &    0.1000  &    0.0000  &    0.9000  &            &    0.1000  &    0.0000  &    1.0000  \\

\hline
           &  \multicolumn{ 3}{c}{(B,E) of PBNOR} &            &  \multicolumn{ 3}{c}{(B,E) of OBNOR} \\

           &      ($F$,$T$) &      ($F$,$F$) &  ($F$,\{$T$,$F$\}) &            &      ($F$,$T$) &      ($F$,$F$) &  ($F$,\{$T$,$F$\}) \\
 \cline{2-4}   \cline{6-8}

         $T$ &    0.7000  &    0.0000  &    0.0000  &            &    0.7000  &    0.0000  &    0.7000  \\

         $F$ &    0.1000  &    1.0000  &    0.9000  &            &    0.1000  &    1.0000  &    0.0000  \\

   \{$T$,$F$\} &    0.1000  &    0.0000  &    0.1000  &            &    0.1000  &    0.0000  &    0.3000  \\

\hline
           &  \multicolumn{ 3}{c}{(B,E) of TBNOR} &            & \multicolumn{ 3}{c}{(B,E) of OCBNOR($\lambda$=0.6)} \\

           &      ($F$,$T$) &      ($F$,$F$) &  ($F$,\{$T$,$F$\}) &            &      ($F$,$T$) &      ($F$,$F$) &  ($F$,\{$T$,$F$\}) \\
 \cline{2-4}   \cline{6-8}

         $T$ &    0.7000  &    0.0000  &    0.3500  &            &    0.7000  &    0.0000  &    0.4200  \\

         $F$ &    0.1000  &    1.0000  &    0.4500  &            &    0.1000  &    1.0000  &    0.3600  \\

   \{$T$,$F$\} &    0.1000  &    0.0000  &    0.2000  &            &    0.1000  &    0.0000  &    0.2200  \\

\hline
\end{tabular} \label{bnorcpdf-b-f} \end{table}

\begin{table}[htp] \centering \caption{The conditional mass distribution for $P(A|B=\{ T,F \}\triangleq \Theta ,E)$ by ImNOR and different BNORs.} \begin{tabular}{rrrrrrrrr} \hline
         $A$ &  \multicolumn{ 3}{c}{($B$,$E$) of ImNOR} &            & \multicolumn{ 3}{c}{($B$,$E$) of LC-BNOR} \\

           &      ($\Theta$,$T$) &      ($\Theta$,$F$) &  ($\Theta$,$\Theta$) &            &      ($\Theta$,$T$) &      ($\Theta$,$F$) &  ($\Theta$,$\Theta$) \\
           \cline{2-4}   \cline{6-8}

         $T$ &    0.7000  &    0.0000  &    0.0000  &            &    0.7000  &    0.0000  &    0.0000  \\

         $F$ &    0.0200  &    0.2000  &    0.0200  &            &    0.0000  &    0.0000  &    0.0000  \\

   \{$T$,$F$\} &    0.0200  &    0.8000  &    0.9800  &            &    0.0000  &    1.0000  &    1.0000  \\

\hline
           &  \multicolumn{ 3}{c}{($B$,$E$) of PBNOR} &            &  \multicolumn{ 3}{c}{($B$,$E$) of OBNOR} \\

           &      ($\Theta$,$T$) &      ($\Theta$,$F$) &  ($\Theta$,$\Theta$) &            &      ($\Theta$,$T$) &      ($\Theta$,$F$) &  ($\Theta$,$\Theta$) \\
 \cline{2-4}   \cline{6-8}

$T$ &    0.7000  &    0.0000  &    0.0000  &            &    0.8800  &    0.6000  &    0.8800  \\

         $F$ &    0.1000  &    1.0000  &    0.9000  &            &    0.0200  &    0.2000  &    0.0000  \\

   \{$T$,$F$\} &    0.1000  &    0.0000  &    0.1000  &            &    0.0200  &    0.2000  &    0.1200  \\

\hline
           &  \multicolumn{ 3}{c}{($B$,$E$) of TBNOR} &            & \multicolumn{ 3}{c}{($B$,$E$) of OCBNOR($\lambda$=0.6)} \\

           &      ($\Theta$,$T$) &      ($\Theta$,$F$) &  ($\Theta$,$\Theta$) &            &      ($\Theta$,$T$) &      ($\Theta$,$F$) &  ($\Theta$,$\Theta$) \\
 \cline{2-4}   \cline{6-8}

         $T$ &    0.7900  &    0.3000  &    0.5450  &            &    0.8080  &    0.3600  &    0.6288  \\

         $F$ &    0.0600  &    0.6000  &    0.2700  &            &    0.0520  &    0.5200  &    0.1872  \\

   \{$T$,$F$\} &    0.0600  &    0.1000  &    0.1850  &            &    0.0520  &    0.1200  &    0.1840  \\

\hline
\end{tabular} \label{bnorcpdf-b-tf} \end{table}

 \begin{center}
 \begin{figure}[!htp] 	\includegraphics[width=0.45\linewidth]{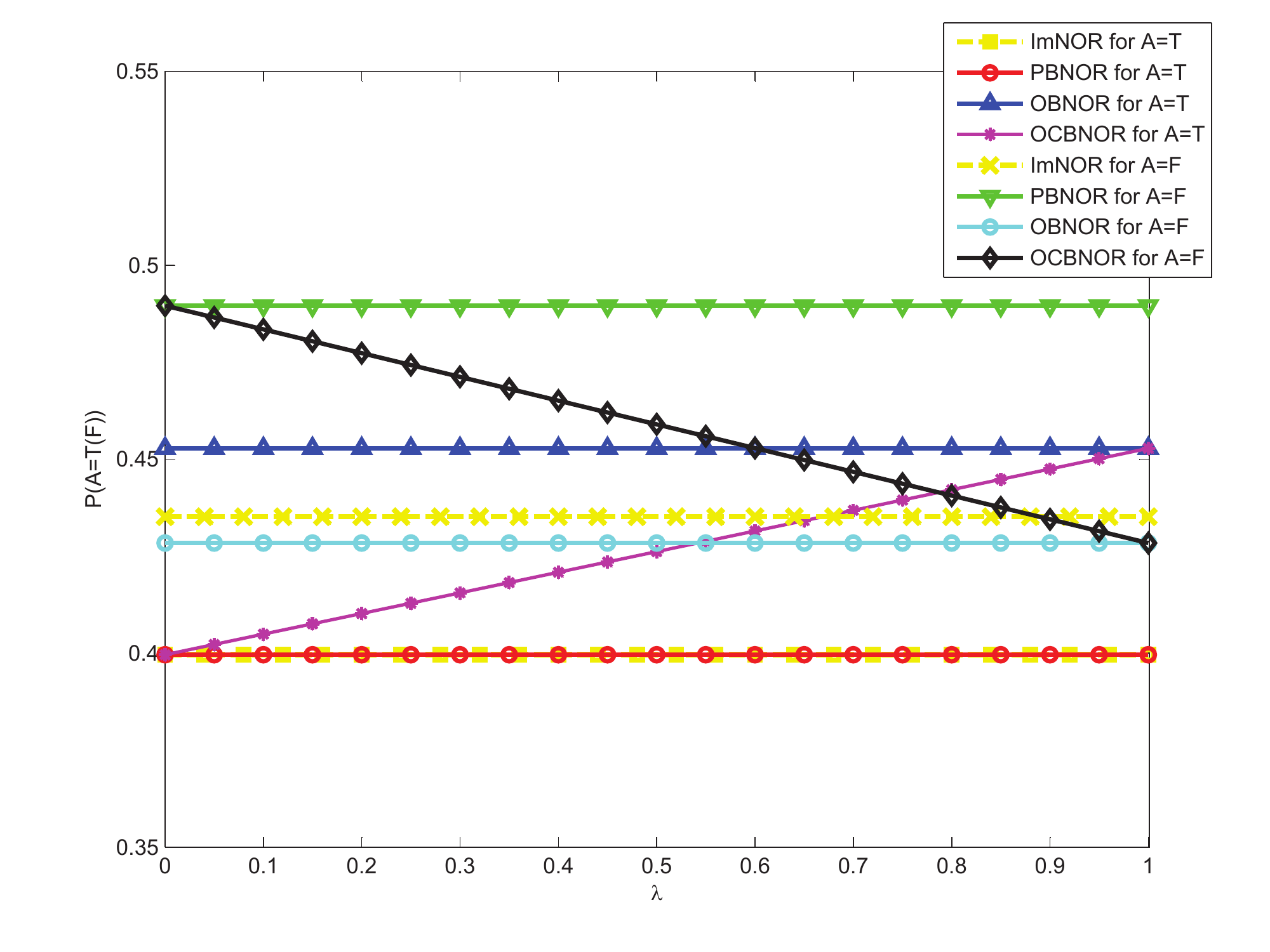} 	\hfill 	\includegraphics[width=.45\linewidth]{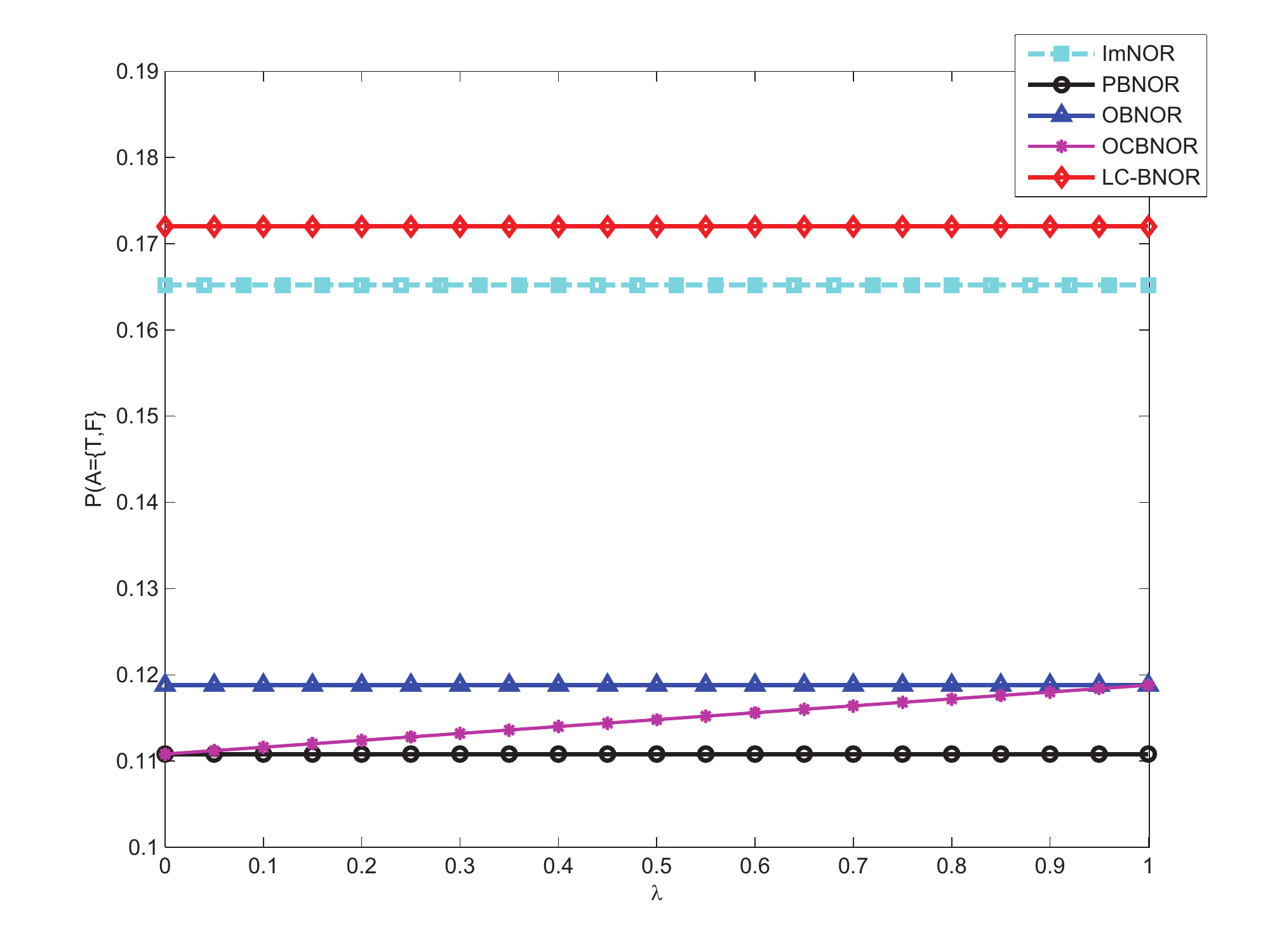} 	\hfill
 \parbox{.45\linewidth}{\centering\small a. $m(A=\{T( F)\})$}
	\hfill
	 \parbox{.45\linewidth}{\centering\small b. $m(A=\{ T,F \})$ }
	\hfill
 \caption{The different BNOR structures for $m(A)$.}
 \label{example0606-child} \end{figure} \end{center}

\iffalse
\begin{table}[ht] \centering \caption{The conditional belief distribution by Fallet's imprecise-Noisy-OR model } \begin{tabular}{rrrrrrr} \hline & $A= T$ & $A= F$ & $A=\{ T, F\}$ \\ \hline $B= T,E= T$ & 0.98 & 0.02 & 0 \\ $B= T,E= F$ & 0.8 & 0.2 & 0 \\ $B= T,E=\{ T, F\}$ & 0.8 & 0.02 & 0.18\\ \hline \end{tabular} \label{simoncpdf} \end{table}
\fi

Marginal mass $m(A)$ can be obtained by ImNOR and different BNORs, and the results are shown in Table~\ref{exampletablechild} and Figure~\ref{example0606-child}. It is shown that OBNOR provides optimistic results while PBNOR produces a pessimistic one.  In order to make a compromise, we can adjust the optimistic coefficient ($\lambda$). Also the contradiction  attitudes of ImNOR can be found here. Thus the BNOR methods are more reasonable and informative.

\begin{table}[htp]
\centering
\caption{The belief mass distribution of the child node $A$.}
\begin{tabular}{rrrrrrr}
\hline

         $A$ &      ImNOR &    LC-BNOR &      PBNOR &      OBNOR &      TBNOR & OCBNOR($\lambda=0.6$) \\
\hline

         $T$ &    0.3996  &    0.3996  &    0.3996  &    0.4528  &    0.4262  &    0.4315  \\

         $F$ &    0.4352  &    0.4284  &    0.4896  &    0.4284  &    0.4590  &    0.4529  \\

   \{$T$,$F$\} &    0.1652  &    0.1720  &    0.1108  &    0.1188  &    0.1148  &    0.1156  \\

\hline \end{tabular} \label{exampletablechild} \end{table}

\begin{center}
 \begin{figure}[!htp] 	\includegraphics[width=0.45\linewidth]{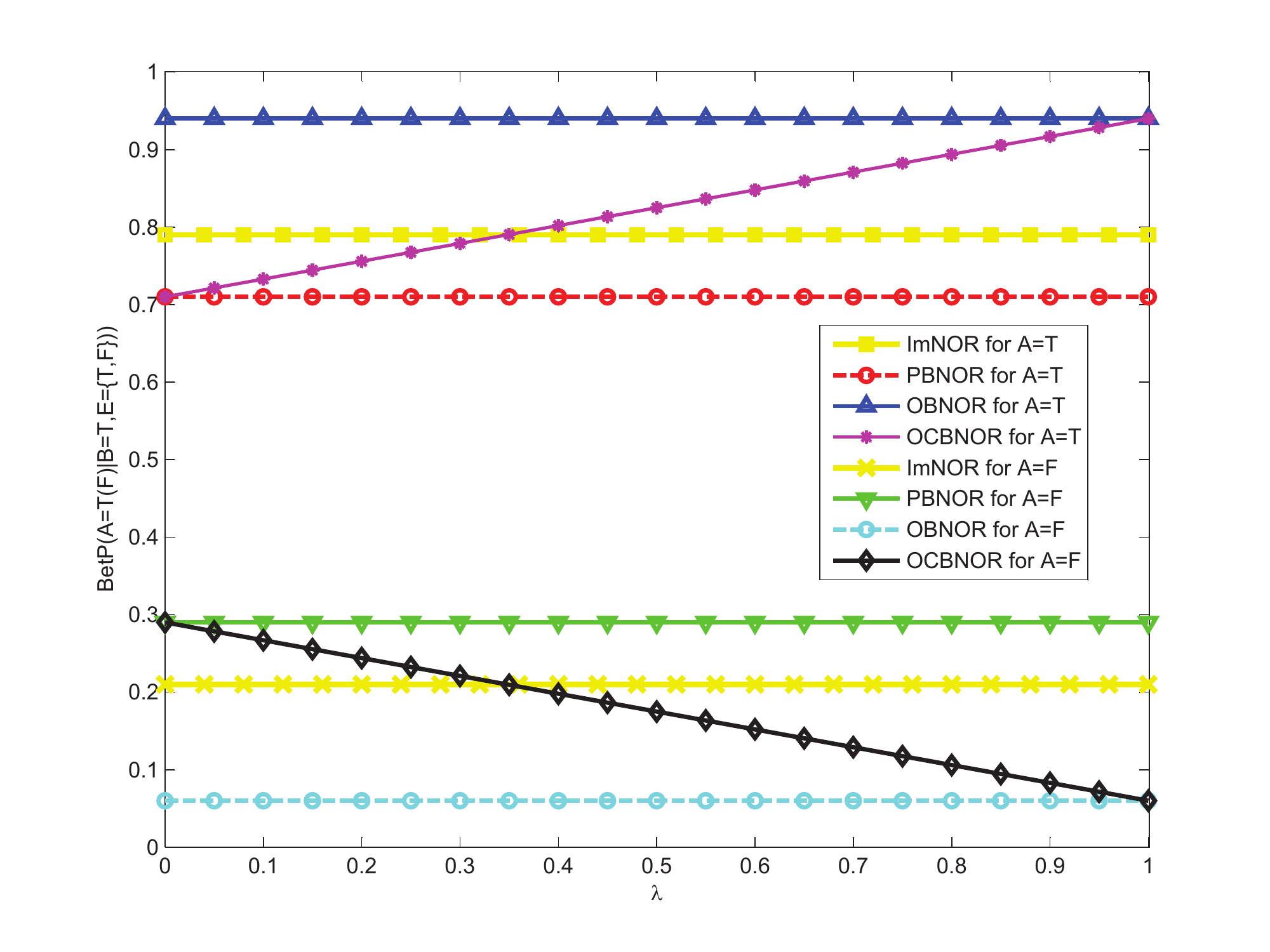} 	\hfill 	\includegraphics[width=.45\linewidth]{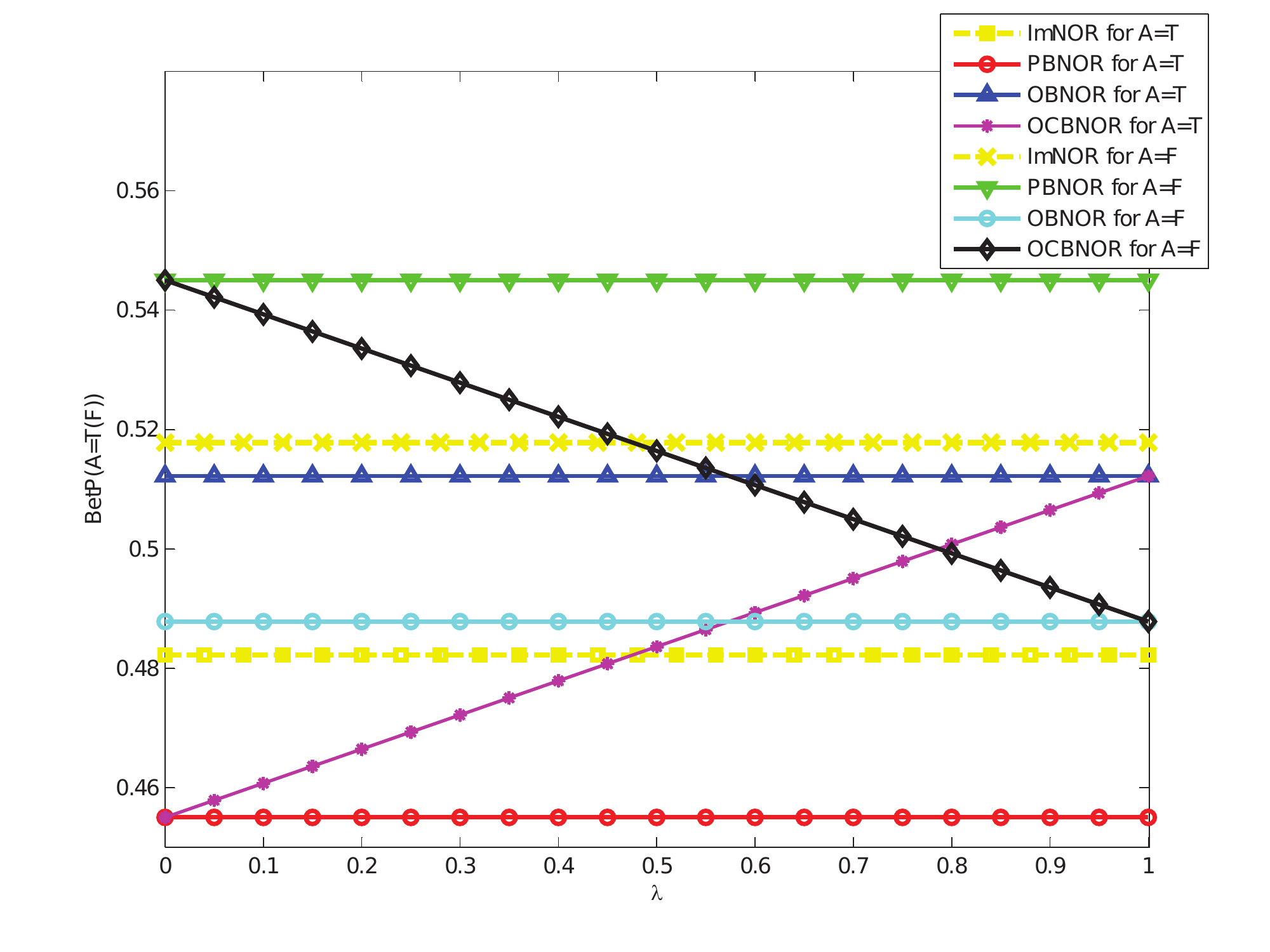} 	\hfill
 \parbox{.45\linewidth}{\centering\small a. $BetP(A|B= T,E=\{T,F\})$}
	\hfill
	 \parbox{.45\linewidth}{\centering\small b. $BetP(A)$ }
	\hfill
 \caption{The (conditional) pignistic probabilities.}
 \label{example0606-pig} \end{figure} \end{center}

Using Eqs.~\eqref{pignistic} and \eqref{pignistic1}, the (conditional) belief mass functions can be transferred
to (conditional) pignistic probabilities. The results of $$BetP(A|B= \{T\},E=\text{\{T,F\}})$$ and $BetP(A)$ are displayed in Figure~\ref{example0606-pig}. It can be seen that BNOR can provide us  abundant information
for decisions with different special requirements. By comparison, ImNOR is more temperate. This is due to the fact that ImNOR assigns more mass values to the vague state $\{ T,F \}$. However, if we want to hold the principle that more belief should be given to the uncertain state, LC-BNOR is a better choice (see Figures~\ref{example0606}-b and \ref{example0606-child}-b).

\section{Network reliability analysis}

%In order to compare methods proposed in this paper, we have devoted
%this section to numerical results.

In this section we will discuss the application of BNOR on the problem of network reliability
analysis. The definition of network reliability
and the traditional Bayesian solution are first recalled. Then the reliability evaluation strategy using BNOR model will be
described in detail.

\subsection{Network reliability and Bayesian network solution }

\label{sub:network reliability}

The network reliability  considered here is two-terminal
reliability, defined as the probability that there is an
operative path between source nodes $n_{1}$ and sink nodes $n_{N}$
in the network. Nodes are assumed to be operational at all times,
and edge failures are assume to be statistically independent.

\begin{center} \begin{figure}[!htp] \includegraphics[width=0.45\linewidth]{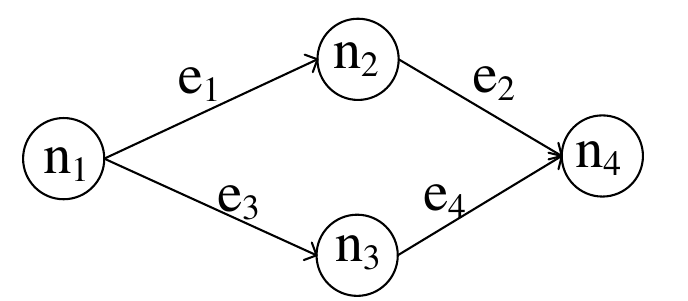}
\hfill
\includegraphics[width=0.45\linewidth]{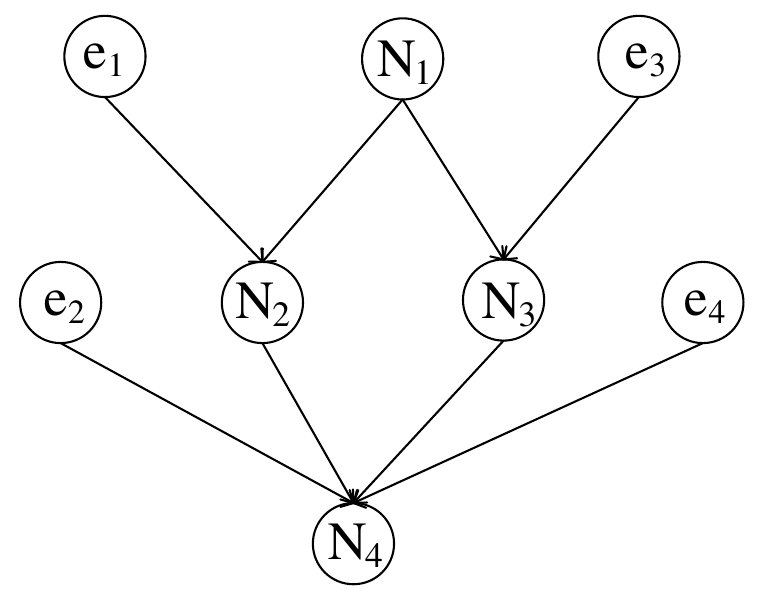}
\hfill
\parbox{.45\linewidth}{\centering\small a. Directed network} 	\hfill 	\parbox{.45\linewidth}{\centering\small b. BN solution}
\hfill
\caption{An example of directed network and its BN solution.}  \label{example1} \end{figure} \end{center}

Let graph $G(N,E)$ represent a network, where $N$ denotes
the set of nodes and $E$ is the set of links. For the network
shown in Figure~\ref{example1}-a, $N=\left\{ n_{1},n_{2},n_{3},n_{4}\right\} $, $E=\left\{ e_{1,}e_{2},e_{3},e_{4}\right\} $.
Define $|N|$ new variables, $N_{i},i=1,2,\cdots,|N|$ , indicating
whether the communication between $n_{1}$ and $n_{i}$ is successful.
Let $N^{+}(n_{i})=\left\{ n_{j}|<n_{j},n_{i}>\in E\right\} $, for every
element of which sets, $n_{j}$, there is a path from $n_{j}$ to
$n_{i}$. And let $E^{+}(n_{i})=\left\{ e_{k}=<n_{j},n_{i}>|n_{j}\in N^{+}(n_{i})\right\} $
denote the sets of edges directed to node $n_{i}$.

\citet{bnsolution} presented a method for reliability evaluating
of networks based on BN. The nodes and edges of BN can be created
according to $G(N,E)$:
\begin{enumerate}
\item The root nodes of BN are $N_{1}$ and the edges in $E$.
\item The non-root nodes of BN are $N_{i},i\neq1$. And
\begin{equation}
Pa(N_{i})=\left\{ N_{j}|n_{j}\in N^{+}(n_{i})\right\} \cup\left\{ e_{k}|e_{k}\in E^{+}(n_{i})\right\}.
\end{equation}
\end{enumerate}
Bayesian networks\textquoteright{} inference algorithms can be evoked
then to calculate the network reliability $P(N_{s})$. The BN framework
for the network described in Figure~\ref{example1}-a can be seen in
Figure~\ref{example1}-b.

\subsection{Reliability evaluation using BNOR}
\begin{center} \begin{figure}[!htp]\centering \includegraphics[width=0.6\linewidth]{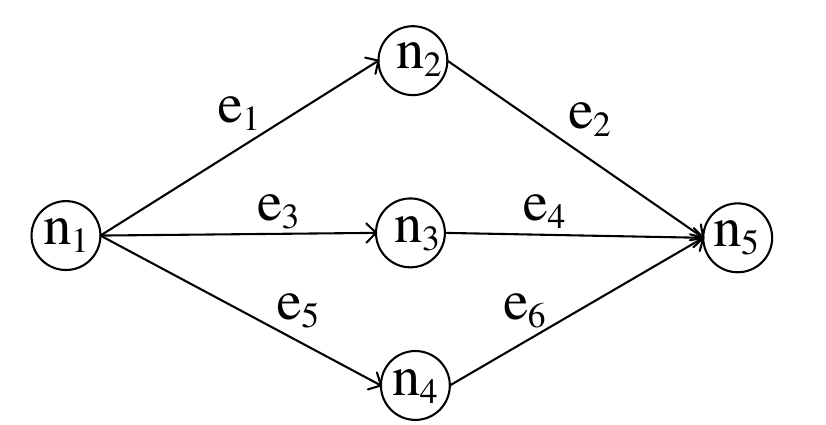}
\caption{The network with 5 nodes.}  \label{network5nodes11}  \end{figure} \end{center}

For the network shown in Figure~\ref{network5nodes11}, its reliability
 can be defined by the probability that there is a working path connected from $n_1$ to $n_5$.
%\[
%P(S= T)=P(\text{there is a working path connected from \ensuremath{n_{1}} to \ensuremath{n_{5}}}).
%\]
The network is operating if the two terminal nodes $n_1$ and $n_5$ are connected by operational
edges. Let $S$ denote the state of the network. It has binary states with  $T$ (working) or $F$ (fail). The values for failure rates of each edge are
\[
\lambda_{e_{1}}=\ensuremath{\lambda_{e_{3}}=\lambda_{e_{5}}=1.5*10^{-3}}~ \mathrm{h^{-1}},\ensuremath{\lambda_{e_{2}}=\lambda_{e_{4}}=\lambda_{e_{6}}=1.8*10^{-3} ~\mathrm{h}^{-1}}.
\]

Consider the mission time $t=200~\mathrm{h}$. The probability distribution
of each edge is given in Table ~$\ref{pdfedge}$.

% latex table generated in R 3.0.0 by xtable 1.7-1 package
% Wed May 29 11:24:53 2013
\begin{table}[ht]
\centering
\caption{The probability distribution of each edge of the network }
\begin{tabular}{rrrrrrr}   \hline  & $e_1$ & $e_2$ & $e_3$ & $e_4$ & $e_5$ & $e_6$ \\
\hline $T$ & 0.8025 & 0.6977 & 0.8025 & 0.6977 & 0.8025 & 0.6977 \\   $F$ & 0.1975 & 0.3023 & 0.1975 & 0.3023 & 0.1975 & 0.3023 \\
% \{T,F\} & 0.0000 & 0.0000 & 0.0000 & 0.0000 & 0.0000 & 0.0000 \\
\hline \end{tabular} \label{pdfedge} \end{table}

\begin{center} \begin{figure}[!htp] \centering
\includegraphics[width=0.6\linewidth]{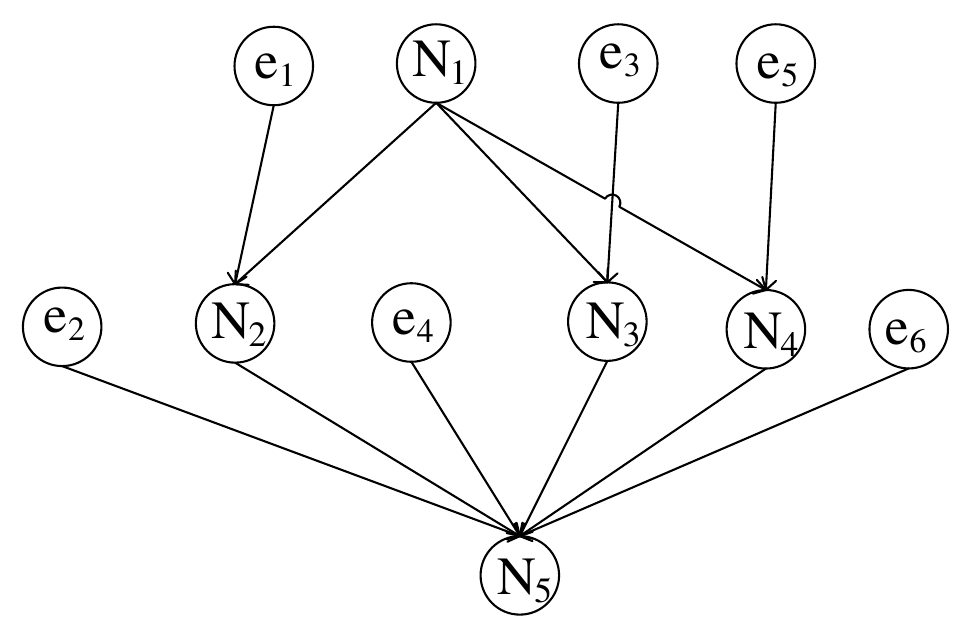} \caption{The BN network solution.} \label{tra_BN}  \end{figure} \end{center}

The traditional Bayesian network  approach is first adopted to calculate the reliability. Using
the method described in Section \ref{sub:network reliability}, we
can create a Bayesian network solution model (see Figure~\ref{tra_BN}). By applying the Junction Tree
(JT) inference algorithm, we can obtain the exact value of the network
reliability $p(S= T)=0.9148$.

\begin{center} \begin{figure}[!htp] \centering \includegraphics[width=0.6\linewidth]{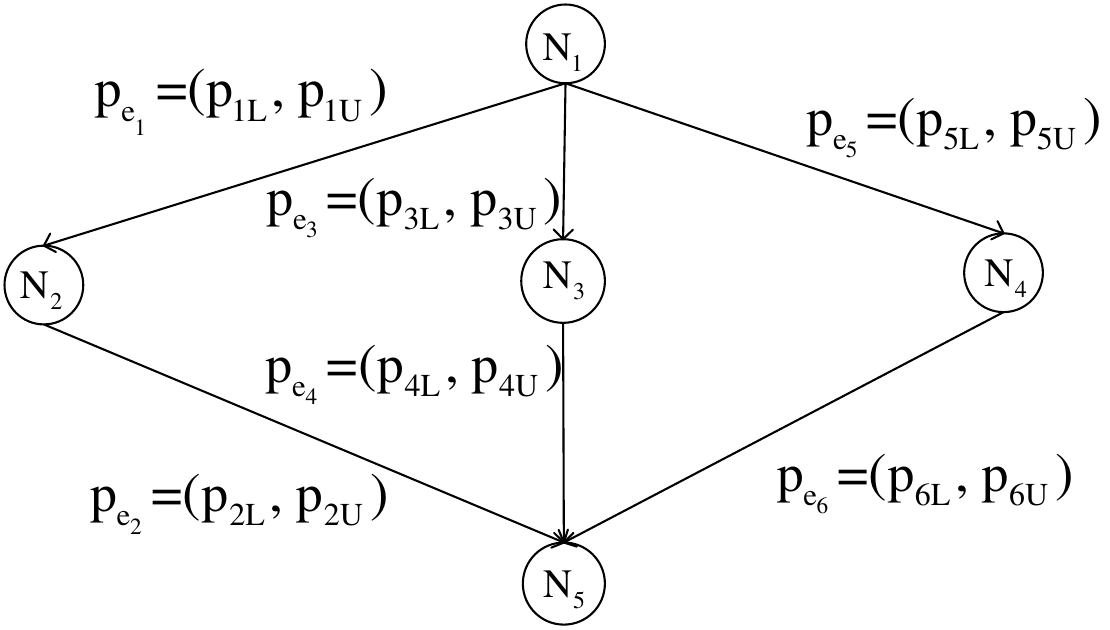}
\caption{The BNOR model for network reliability.} \label{Belief_EV}  \end{figure} \end{center}

In the next two subsections, different BNOR structures (PBNOR, OBNOR, TBNOR, OCBNOR and LC-BNOR) will be applied to estimate
the network reliability. %The evaluation
%results will be compared with those by the use of ImNOR.
The BNOR model for this network is shown in Figure~\ref{Belief_EV}. The fail of edge $e_{i}$
can be regarded as an inhibition. For example, in the network displayed
in Figure~\ref{network5nodes11}, even if $N_2$ is in the working state $T$, $N_5$  may not in state $T$ due
to the fail of $e_{2}$. The inhibition  can be  measured by link probabilities $p_{e_i}, i = 1,2,\cdots,5$.

\subsection{BNOR model, case with no epistemic uncertainty}

If there is no epistemic uncertainty, neither on the variable state or on link probabilities, {\em i.e.,}
$$m(N_1=\left\{  T,F \right\} )=0,
$$
and
$$
p_{iL}=p_{iU}=p(e_i=T), i=1,2,\cdots,6,
$$
the same results can be obtained by the use of all BNORs and ImNOR:

\[
m(S= \{T\})=0.9148,~m(S= \{F\})=0.0852,~m(S=\{ T,F \})=0.
\]

It can be seen that, if there is no epistemic
uncertainty introduced, the reliability estimation results previously obtained
by traditional BN method can be recovered. This fact indicates that BNOR is a general extension of Noisy-OR structure
in the framework of belief functions, and it is entirely compatible with the probabilistic solution. The $Bel$ and $Pl$ measures can be obtained by Eqs.~\eqref{bel} and \eqref{pl} respectively. The following results can be obtained:

\begin{equation}
  Bel(S= \{T\})=0.9148,~p(S= T)=0.9148,~Pl(S=\{T\})=0.9148.
\end{equation}

\iffalse
\begin{center} \begin{figure}[!htp] 	\includegraphics[width=0.45\linewidth]{1.pdf} 	\hfill 	\includegraphics[width=.45\linewidth]{2.pdf} 	\hfill       \includegraphics[width=.45\linewidth]{3.pdf}     \hfill     	\includegraphics[width=0.45\linewidth]{4.pdf} 	\hfill 	\includegraphics[width=.45\linewidth]{5.pdf} 	\hfill       \includegraphics[width=.45\linewidth]{6.pdf}     \hfill     	\includegraphics[width=0.45\linewidth]{7.pdf} 	\hfill 	\includegraphics[width=.45\linewidth]{8.pdf} 	\hfill       \includegraphics[width=.45\linewidth]{9.pdf}     \hfill 	\caption{Basic belief assignments and network reliability with no epistemic uncertainty} \label{nouncertainty} \end{figure} \end{center}
\fi

\subsection{BNOR model, case with an epistemic uncertainty}

In this test, the case with some epistemic uncertainty  on
 link probabilities is considered.
 %Let the belief mass assigned to $\{T,F\}$ be $$m(e_{1}=\{ T,F \})=0.1, ~~m(e_{2}=\{ T,F \})=0.1,$$  on the  inhibition
%the link probabilities of $N_{2}$ and $N_{1}$,  is introduced.
The lower and upper probability bounds of link probabilities  $p_{e_i}$ are listed in Table~\ref{inhibition table}. The results
by  five BNORs and ImNOR are illustrated in Table~\ref{networktablechild}.
\begin{table}[htp]
\centering
\caption{The probability intervals of the inhibition parameters }
\begin{tabular}{rrrrrrr}
\hline

          &      $p_{e_1}$ &    $p_{e_2}$ &      $p_{e_3}$ &     $p_{e_4}$ &    $p_{e_5}$ & $p_{e_6}$ \\
\hline

        $ P_{iL}$ &     0.7525  &  0.6477   &  0.8025   &  0.6977  &   0.8025  &   0.6977 \\

         $P_{iU}$ &    0.8525  &   0.7477   &  0.8025   &  0.6977  &   0.8025   &  0.6977 \\

\hline \end{tabular} \label{inhibition table} \end{table}

\begin{center}
 \begin{figure}[!htp] 	\includegraphics[width=0.45\linewidth]{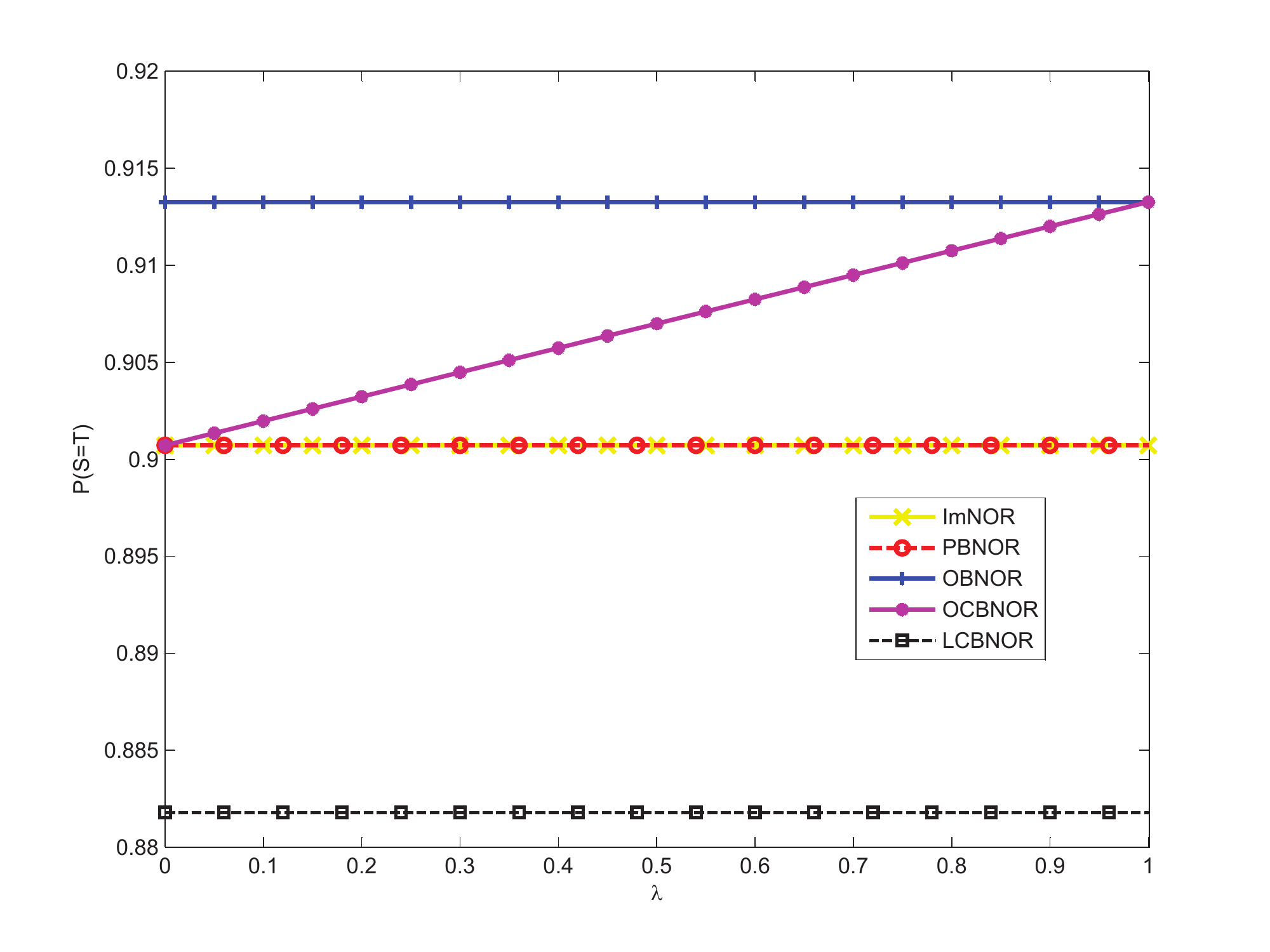} 	\hfill 	\includegraphics[width=.45\linewidth]{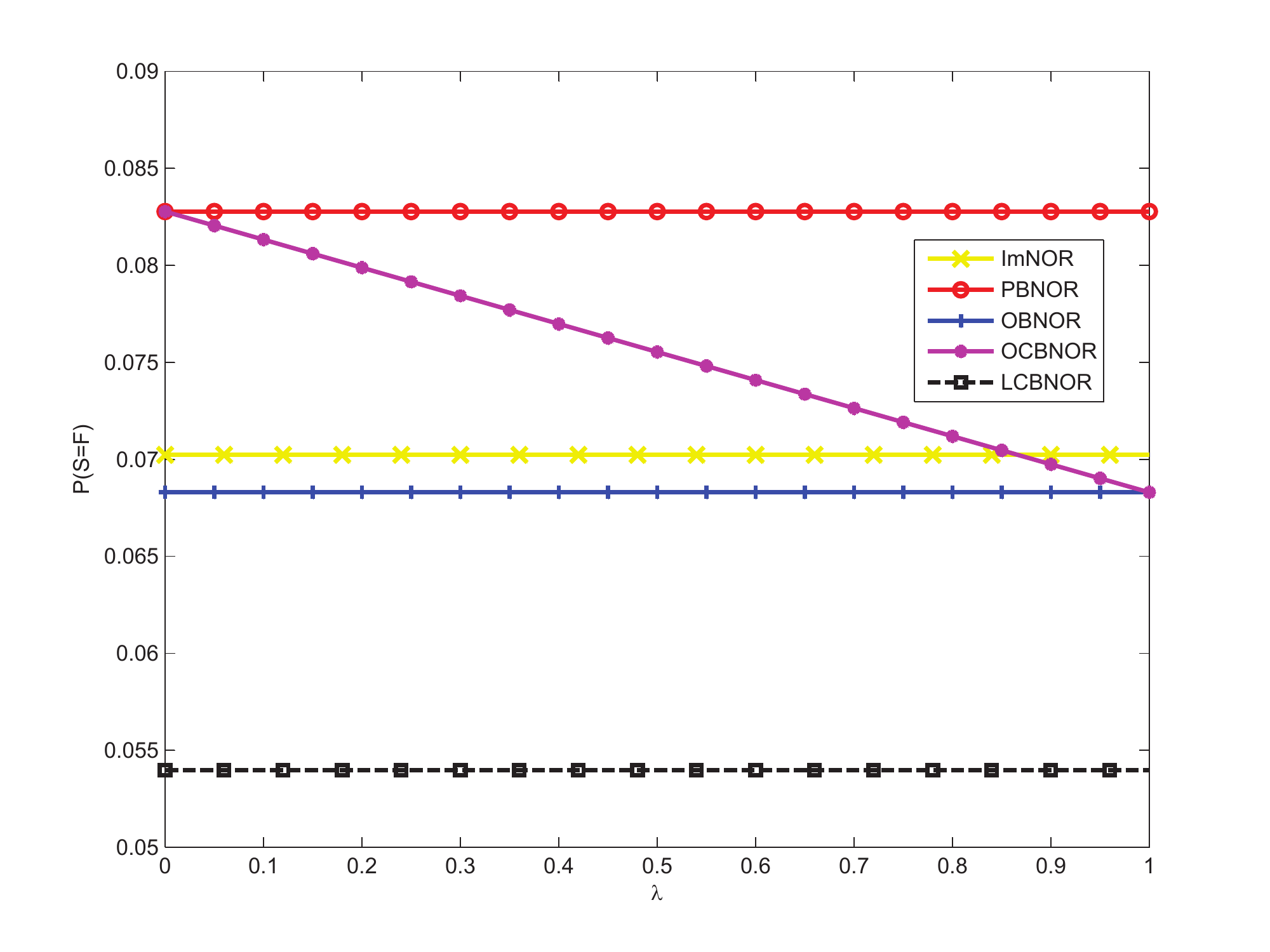} 	\hfill
 \parbox{.45\linewidth}{\centering\small a. $m(S= \{T\})$}
	\hfill
	 \parbox{.45\linewidth}{\centering\small b. $m(S= \{F\})$ }
	\hfill
\includegraphics[width=0.45\linewidth]{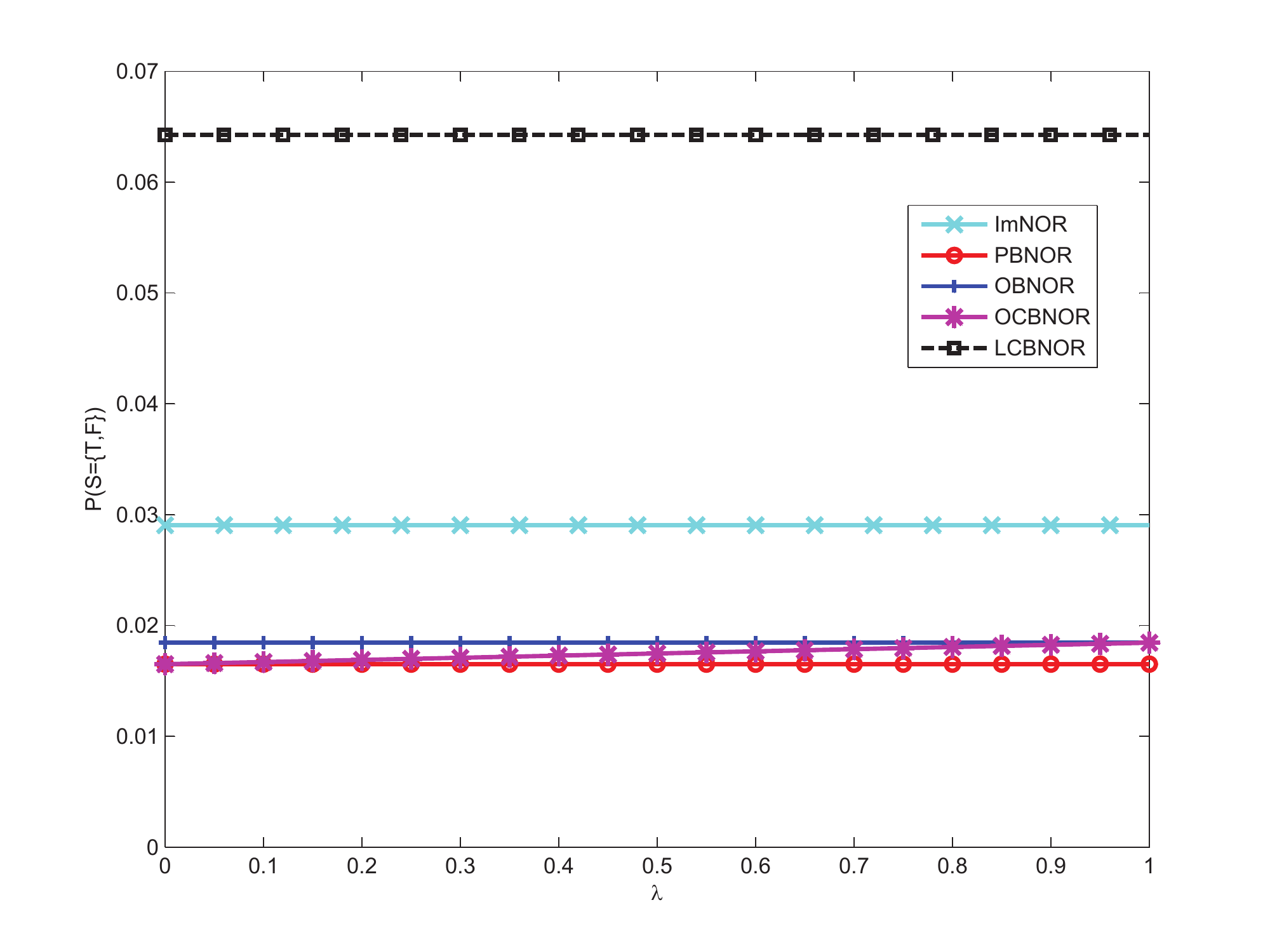} 	\hfill 	\includegraphics[width=.45\linewidth]{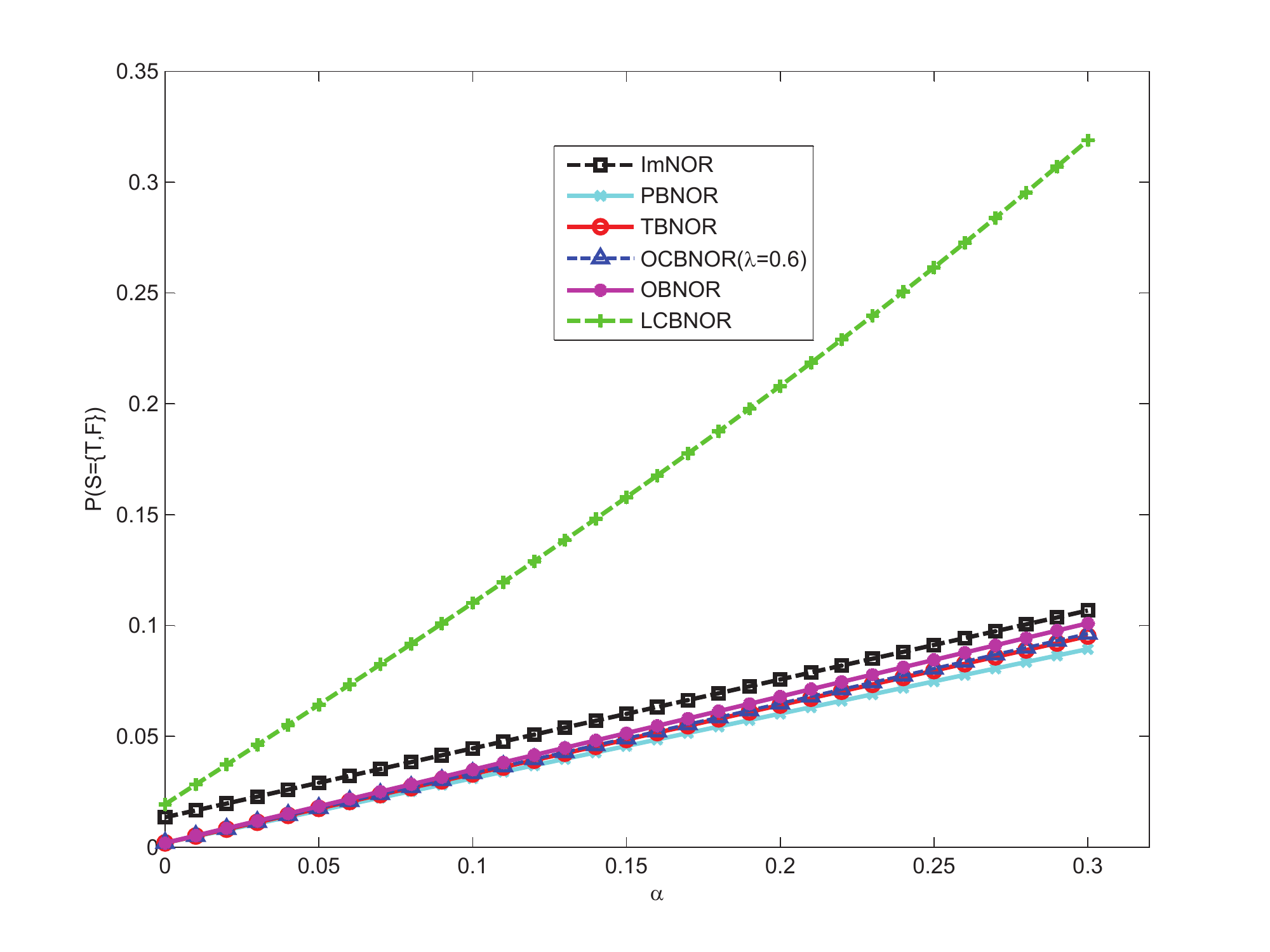} 	\hfill
 \parbox{.45\linewidth}{\centering\small c. $m(S=\{ T,F \})$ varying with $\lambda$}
	\hfill
	 \parbox{.45\linewidth}{\centering\small d. $m(S=\{ T,F \})$ varying with $\alpha$ }
 \caption{The different BNOR structures for $m(S)$.}
 \label{examplenetwork-child} \end{figure} \end{center}

In Figures~\ref{examplenetwork-child}-a and \ref{examplenetwork-child}-b, the  results of  the evaluation of the network system  varying with
 the optimism coefficient $\lambda$ are demonstrated.
It can be found that when $\lambda$ grows from 0 up to 1, the mass on ``the system is working" ({\em i.e.,} $m(S= \{T\})$) increases. On the
 contrary, $m(S= \{F\})$ decreases.  This meets with our common sense of  optimistic and pessimistic decisions. However, ImNOR provides
  opposite attitudes towards  $m(S= \{T\})$ and $m(S= \{F\})$. The performance of ImNOR is similar to PBNOR for calculating $m(S=\{T\})$, while
  similar to OBNOR for calculating $m(S=\{F\})$.

Figure~\ref{examplenetwork-child}-c illustrates that the mass assigned to the uncertain state $\{T,F\}$ by LC-BNOR is larger than that
by the other models. This is due to the principle LC-BNOR upholds is that  one should never give more support than justified
to any subset of the discernment frame.

Let the interval of link probability $p_{e_1}$ be $[0.7525,0.8525]$ and the corresponding interval of $p_{e_2}$ be $[0.6977,0.6977]-[\alpha,-\alpha]$.
The length of the interval, $2\alpha$, reflects the degree of uncertainty to some extend.
Figure~\ref{examplenetwork-child}-d depicts the mass assigned to the
 uncertain state $m(S=\{T,F\})$ varying with
 $\alpha$. It  can be seen that the uncertainty on system's state increases with the increasing of $\alpha$.
LC-BNOR is the most sensitive  to uncertainty variation among the six NOR models.

 In this case, the credibility and plausibility measures on $S = \{T\}$ are not
the same, and they bound the network reliability.
For example, if we use OCBNOR ($\lambda=0.6$), from Table~\ref{networktablechild}
 we can see the belief mass assignment of the network state $S$:
 \begin{equation}\label{bba_network}
   m(S= \{T\})=0.9082, ~m(S= \{F\})=0.0741,~ m(S=\{ T,F \})=0.0177.
 \end{equation}
The following results can be obtained:
 \begin{equation}\label{bba_network_bel_pl_t}
   Bel(S= \{T\})=0.9082, ~p(S= T)=0.9148, ~Pl(S= \{T\})=0.9259.
 \end{equation}
%  \begin{equation}\label{bba_network_bel_pl_f}
%   Bel(S= F)=0.0741, Pl(S= F)=0.0918.
% \end{equation}
Further decisions can be made based on these uncertainty knowledge.
\begin{table}[htp]
\centering
\caption{The belief mass distribution of the network system.}
\begin{tabular}{rrrrrrr}
\hline

         S &      ImNOR &    LC-BNOR &      PBNOR &      OBNOR &      TBNOR & OCBNOR($\lambda=0.6$) \\
\hline

         $T$ &    0.9007  &    0.8818  &    0.9007  &    0.9133  &    0.9070  &    0.9082  \\

         $F$ &    0.0702  &    0.0540  &    0.0828  &    0.0683  &    0.0755  &    0.0741  \\

   \{$T$,$F$\} &    0.0291  &    0.0642  &    0.0165  &    0.0184  &    0.0175  &    0.0177  \\

\hline \end{tabular} \label{networktablechild} \end{table}

\begin{center}
 \begin{figure}[!htp] 	\includegraphics[width=0.45\linewidth]{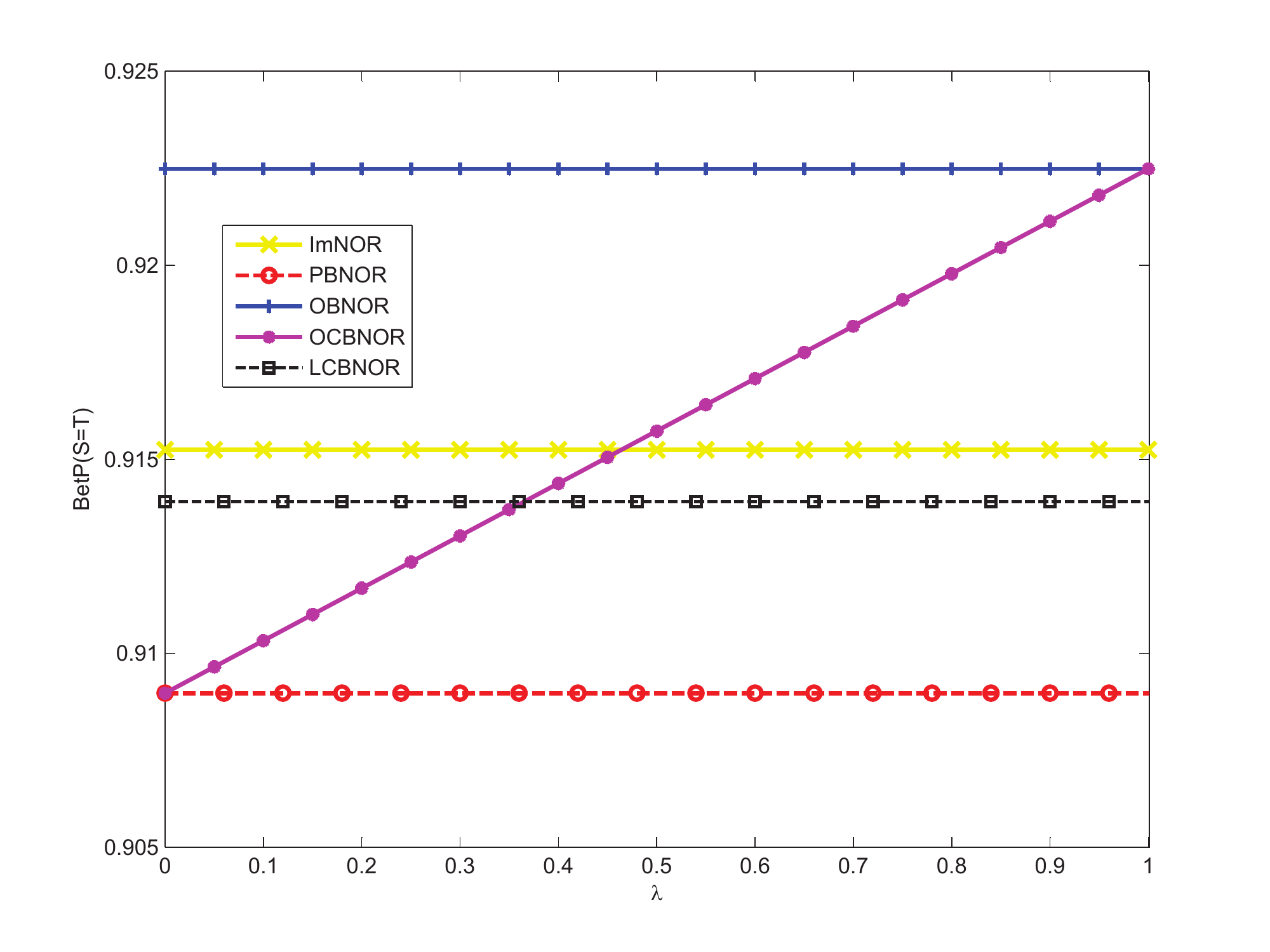} 	\hfill 	\includegraphics[width=.45\linewidth]{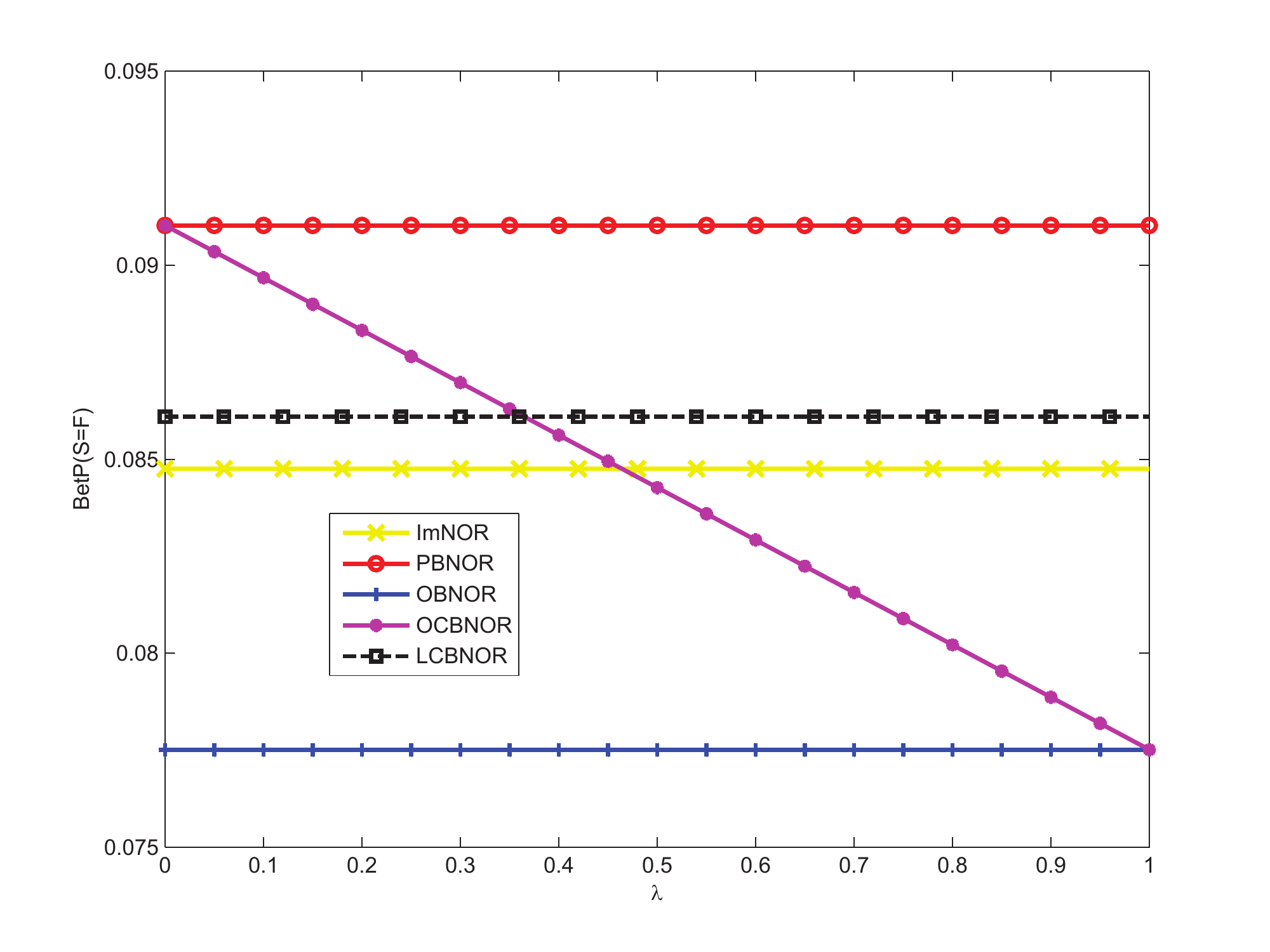} 	\hfill
 \parbox{.45\linewidth}{\centering\small a. $BetP(S= T)$}
	\hfill
	 \parbox{.45\linewidth}{\centering\small b. $BetP(S= F)$ }
\caption{The different BNOR structures for $BetP(S)$.}
 \label{examplenetwork-child-pig} \end{figure} \end{center}

The information on  credal level can be transformed to the pignistic level  by Eq.~\eqref{pignistic} where the
decisions can be made more easily. The probability distributions by different models are listed in Table ~\ref{Pignistictab}. As it
can be observed in Figure~\ref{examplenetwork-child-pig}, with the increasing  of $\lambda$, the pignistic probability for
$S= T$ increases, on the other hand decreases for $S= F$. For $S= T$, OBNOR gives a upper bound  and PBNOR gives a lower bound. While for $S= F$,
OBNOR gives a lower bound and PBNOR gives a upper bound.
This is in accordance with the optimistic and pessimistic principles. However, the attitude of ImNOR is ambiguous.

\begin{table}[htp]
\centering
\caption{The probability distribution of the network system.}
\begin{tabular}{rrrrrrr}
\hline

         S &      ImNOR &    LC-BNOR &      PBNOR &      OBNOR &      TBNOR & OCBNOR($\lambda=0.6$) \\
\hline

          $T$ &     0.9152 &     0.9139 &     0.9090 &     0.9225 &     0.9157  &      0.9171 \\

          $F$ &     0.0848 &     0.0861 &      0.0910 &     0.0775 &     0.0843&      0.0829 \\

\hline \end{tabular} \label{Pignistictab} \end{table}

\subsection{Discussion}

From the experimental results, it can be concluded  that when there is no epistemic uncertainty in the
network, BNOR degrades to the traditional Bayesian method. However,
if there indeed exists imperfect knowledge, BNOR model can provide
us more information in the form of lower bounds ($Bel$) and upper
bounds ($Pl$). A pessimistic
decision can be made according to the $Bel$ value, which provides
us the worst value of the network reliability. On the contrary, $Pl$
is the maximum degree of belief that can support the connectivity
of the network, from which an optimistic decision can be made. These
measures are of great value when we have to make a compromise between risks
and costs. Also these knowledge on the credal level can be
transformed to the pignistic level through Smets method. Then common  principles in decision theory
such as the minimal expected cost (risk) criterion can be evoked.

%One advantage of BN is that backward inference can be implemented,
%that is to say, the probability of causes could be calculated if the
%results are available. This is very important to the importance analysis
%in the system. This results show that BNOR is also more informative in the importance
%analysis than the existing approach.

\section{Conclusion}
%the theory of belief functions  is adopted to represent the
%uncertain information in the network. %, which may be in the form of
%intervals. The length of the interval measures the degree of uncertainty. Moreover,
%there may also be uncertainty  on  the  state  of Boolean
%parent variables.
In this paper, the BNOR model under the framework of belief functions, as an extension of the traditional NOR model, is developed
to express  several kinds of uncertainty  in the
independent causal interactions.
%BNOR in the form of  imprecise intervals and that on the
%variable state at the same time.
It is proved that BNOR can be implemented to propagate uncertain information through employing the Bayesian networks tools.
In practice, BNOR is more flexible than existing NOR models because it offers a general Bayesian framework
that allows us to adopt the exact inference algorithm in their original
form without modification, and then by adjusting necessary parameters to obtain results which satisfy the special requirements of engineers.
Finally, the application  on the reliability evaluation problem of networked systems demonstrates the effectiveness of BNOR model.

BNOR is applicable to systems with binary variables. An extension of BNOR suitable to multi-value variables
should be regarded as a future development. Furthermore, the uncertainty on link probabilities
may not be given in the form of intervals in practice. How to  take advantage of the independence of causal interactions
with different types of uncertain knowledge is another
considerable problem to be investigated. This paper mainly focuses on theoretical work.
Applications of BNOR on more complex situations will be considered in the future.

%In real applications, the uncertain
%information varies in forms which may not be restricted in intervals. Thus how to use
%BNOR to deal with different types of uncertain knowledge is an
%interesting problem to investigate in the future. Furthermore, due to the fact that the structure of
%the network varies with time, the networked systems are often
%dynamic. Thus the problems with time dependence should be considered.

\section*{Acknowledgments.} This work was supported by the National
Natural Science Foundation of China (Nos.61135001, 61403310). The study of the
first author in France was supported by the China Scholarship Council.

%\bibliographystyle{elsarticle-num-names}
%\bibliographystyle{IEEEtran}
%\bibliographystyle{plain}
%\addcontentsline{toc}{section}{\refname}\bibliography{paperlist}

\iffalse
\section*{References}
\noindent
References are to be listed in the order cited in the text. Use
the style shown in the following examples. They are to be cited
in the text after punctuation marks, using superscripts without
brackets.  For journal names, use the standard abbreviations.
Typeset references in 9 pt Times Roman.

\fi
\end{document}